\documentclass[letterpaper, 10 pt, conference]{ieeeconf}  

\IEEEoverridecommandlockouts                              

\newcommand{\citep}[1]{\cite{#1}}
\usepackage{graphics} 
\usepackage{epsfig} 
\usepackage{mathptmx} 
\usepackage{times} 
\usepackage{amsmath} 
\usepackage{amssymb}  
\usepackage{hyperref}  
\usepackage{graphicx}
 \usepackage{pdfpages}
\usepackage[tight,footnotesize]{subfigure}

\title{\LARGE \bf
Advanced Mapping Robot and High-Resolution Dataset
}

\author{Hongyu Chen$^{1}$, Zhijie Yang, Xiting Zhao, Guangyuan Weng, Haochuan Wan, Jianwen Luo, \\Xiaoya Ye, Zehao Zhao, Zhenpeng He, Yongxia Shen, and S\"oren  Schwertfeger$^{1}$
\thanks{$^{1}$All authors are with the School of Information Science and Technology, 
	ShanghaiTech University, China.
	{\tt\small <chenhy3, soerensch>@shanghaitech.edu.cn}
}%
}

\begin{document}

\maketitle
\thispagestyle{empty}
\pagestyle{empty}
\begin{abstract}
This paper presents a fully hardware synchronized mapping robot with support for a hardware synchronized external tracking system, for super-precise timing and localization. Nine high-resolution cameras and two 32-beam 3D Lidars were used along with a professional, static 3D scanner for ground truth map collection. With all the sensors calibrated on the mapping robot, three datasets are collected to evaluate the performance of mapping algorithms within a room and between rooms. Based on these datasets we generate maps and trajectory data, which is then fed into evaluation algorithms. We provide the datasets for download and the mapping and evaluation procedures are made in a very easily reproducible manner for maximum comparability. We have also conducted a survey on available robotics-related datasets and compiled a big table with those datasets and a number of properties of them. 

\textbf{Keywords:} Mobile Robot, Sensor Synchronization, Sensor Calibration, Robotic Datasets, Simultaneous Localization and Mapping (SLAM)
\end{abstract}

\section{INTRODUCTION}
\label{sec:introduction}
Localization and mapping are essential robotic tasks and are often solved together in a Simultaneous Localization and Mapping (SLAM) system \cite{cadena2016past}. SLAM is very much depending on the sensor data. Robotics is profiting from quickly developing sensor technology, for Lidars thanks to industry engagement in autonomous driving and for cameras thanks to digital consumer products. Research on SLAM requires the comparison of the results of the algorithms, the localization (path) and the map, with other SLAM algorithms, in order to show the capabilities of the proposed method.

In order to evaluate the precision and accuracy of the robot's trajectory and map, ground truth information is critically important, however, sometimes it is not a trivial task to get ground truth information.

Many approaches have been employed to measure the quality of SLAM systems. Generally, this can be divided into three categories. The first category is consisting of algorithms which utilize the ground truth robots paths. It is often assumed that good localization results are equivalent to good maps. In \cite{WulfGroundTruthEval2007} and \cite{Kuemmerle2009onMeasuring} the ground truth paths are compared with the paths estimated by the SLAM algorithms. In \cite{TUMRGBD}, it is proposed to use the relative pose error and absolute trajectory error to evaluate the performance of SLAM systems. A metric for measuring the error of the manually corrected trajectory of datasets is also available to the public in \cite{Burgard-SLAMcompGraphOfRelations-IROS09}. Recently, \cite{zhang2018tutorial} has provided a tutorial and software for quantitative trajectory evaluation.  

Another category of evaluation algorithms is not using ground truth paths but the maps created by the mapping system for evaluation. Image similarity methods \cite{MapQuality-RoboCup08} and pixel-level feature detectors \cite{PellenzMappingAndMapScoring2008}, \cite{Lakaemper2009VirtualScans} can be adopted to evaluate the quality of maps created by their algorithms. However, these methods have their own limitations because maps often have errors like structures appearing more than once due to localization errors. In \cite{Schwertfeger2010Fiducials,SSRR11-Schwertfeger-Fiducials}, high-level features like barrels for evaluation of maps both in 2D and 3D maps are applied \cite{Schwertfeger2015Fiducials3D}.
The third group is to utilize the topology of the maps and use the matches for comparison, such as \cite{Schwertfeger2015_Topo_AuRo}. There are also evaluation methods that don't rely on ground truth data. In \cite{Newmann-MarkovFields-MapQuality08}  suspicious and plausible arrangements of planes in 3D scans are detected and the map is evaluated accordingly.

\begin{figure}[t]
	\centering 
	\includegraphics[width=0.9\linewidth]{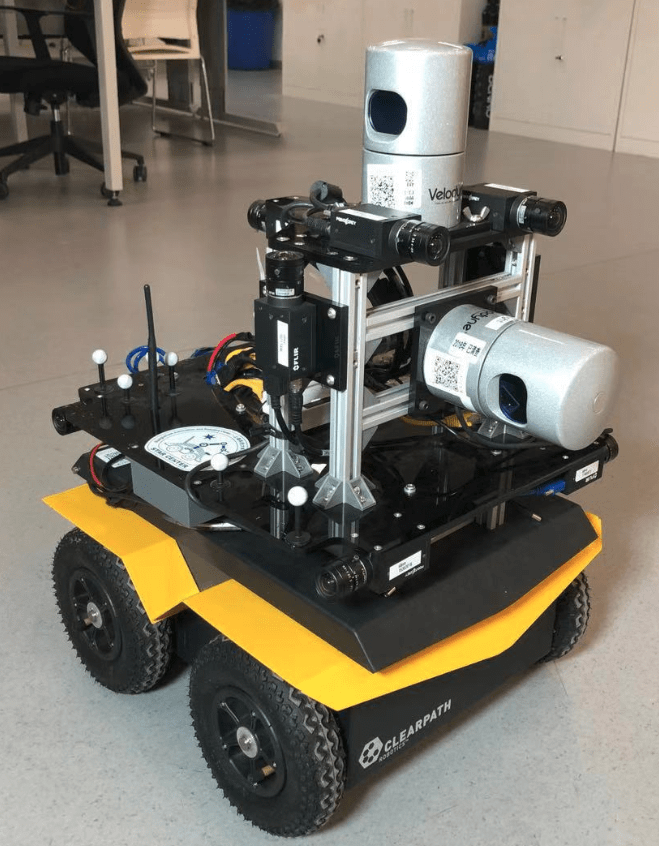}  
	\caption{The MARS Mapper robot with sensors, that is used in this paper. }
	\label{fig:robot_platform}   
\end{figure}

However the evaluation is done, for comparing different SLAM algorithms it is essential that they use the same data. If the algorithms themselves use different data (e.g. Lidar vs. camera data), at least this data should have been collected at the same time with the same robot. Thus collecting mapping datasets and then using those for the evaluation of the SLAM algorithms is an excellent way to characterize the capabilities of the different mapping systems. In the next section of this paper a survey on publicly available mapping datasets is presented. The key finding there is, that there is a lack of high-resolution, heterogeneous mapping datasets that make use of the latest sensor technology. But those are essential to develop and compare now the SLAM systems that may be used in future robotic systems, which may well provide such rich data for a much lower cost compared to todays systems. 
 
Therefore, in this paper, we present an advanced mapping system with high-resolution sensors that are hardware-synchronized to an external tracking system to collect data for benchmarking datasets: the Mars Mapper robot (MARS is the acronym of our Mobile Autonomous Robotic Systems Lab). Preliminary results of this work have been previously presented in the ICRA 2019 Workshop on Dataset Generation and Benchmarking of SLAM Algorithms for Robotics and VR/AR \cite{chen2019towards}. We believe this approach is a valuable supplement to SLAM evaluation using simulations, because it ensures real sensor noise and real locomotion, vibrations and other factors that are difficult to accurately simulate. Using the tracking system we gather ground truth localization information. But we believe that it is also important to evaluate the mapping performance, especially for visual SLAM. So we also collect ground truth map information using a professional, static 3D scanner (Faro Focus 3D). From the 3D ground truth map we also extract a 2D ground truth map. 

 \begin{figure}[t]
  \centering
  \includegraphics[width=\linewidth]{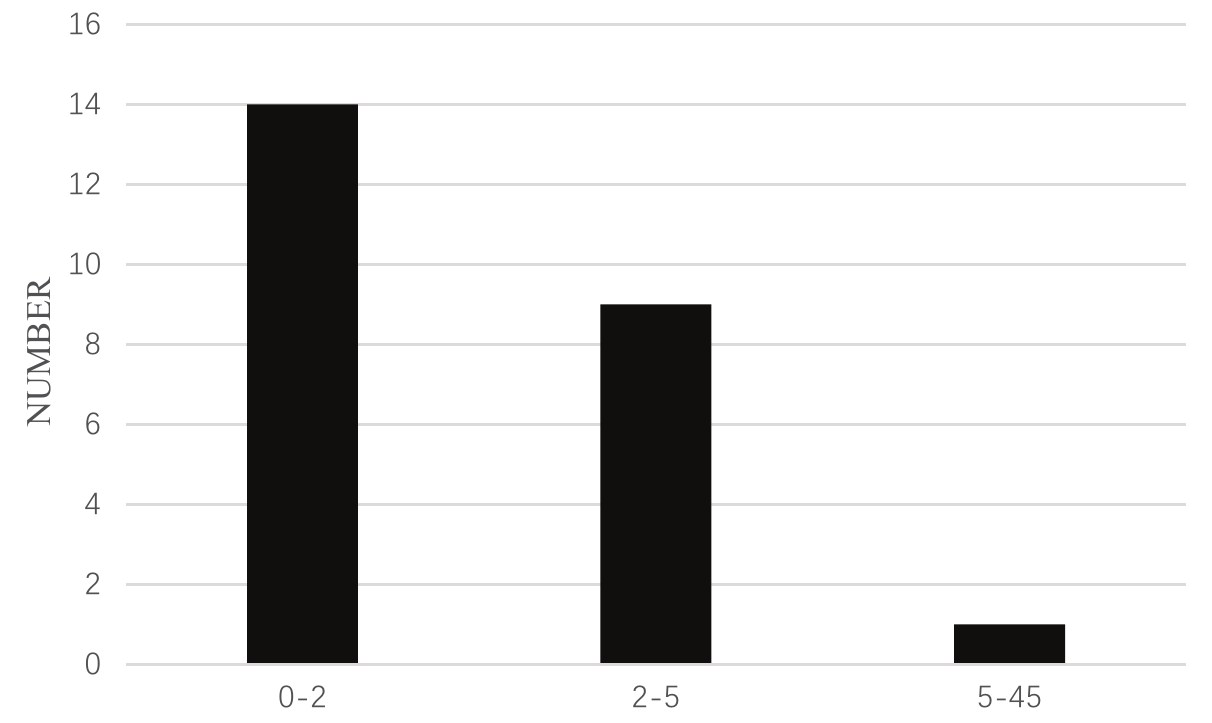}  
  \caption{ Total Mega pixels in SLAM Datasets  which contain RGB cameras }
  \label{fig:megapixels}
\end{figure}

We provide three indoor datasets. One purely within the tracking system, especially also for evaluating mapping performance. Additionally, we have two longer, very similar datasets, that start and end in the tracking system. One can do loop closing at the end, the other cannot (no overlapping sensor data). Those datasets allow a thorough comparison of SLAM algorithms with different settings, sensor combinations, resolutions and framerates. We show some example SLAM evaluations using our datasets, evaluating their loop-closing and scan-matching performance.

We provide ROS launch files to automatically generate the maps from the datasets and to evaluate them by comparison to ground truth. Our results are thus fully reproducible. 

The key contributions of this paper are thus:
\begin{itemize}
	\item A comprehensive survey on mapping datasets, which is condensed to a big table comparing select parameters of the datasets.
	\item Presentation of a hardware-synchronized advanced mapping robot with external tracking.
	\item Generation of three datasets for mapping and SLAM evaluation.
	\item Providing reproducible evaluations of standard mapping software based on the datasets.
\end{itemize}


The remainder of the paper is organized as follows:

Section \ref{sec:dataset_survey} presents the survey on mapping datasets. Then 
Section \ref{sec:mapper} provides a short survey on mapping robots and describes our mapping robot hardware.
Sections \ref{sec:sync} and \ref{sec:calibration} detail the sensor synchronization and calibration approaches for the MARS mapping robot.
In Section \ref{sec:datasets} we describe the three datasets we provide. 
In Section \ref{sec:eval} we run a selection of SLAM algorithms with our datasets and evaluate their performance. The conclusions then follow in Section \ref{sec:conclusions}.

  \begin{figure}[t]
    \centering
    
    \includegraphics[width=\linewidth]{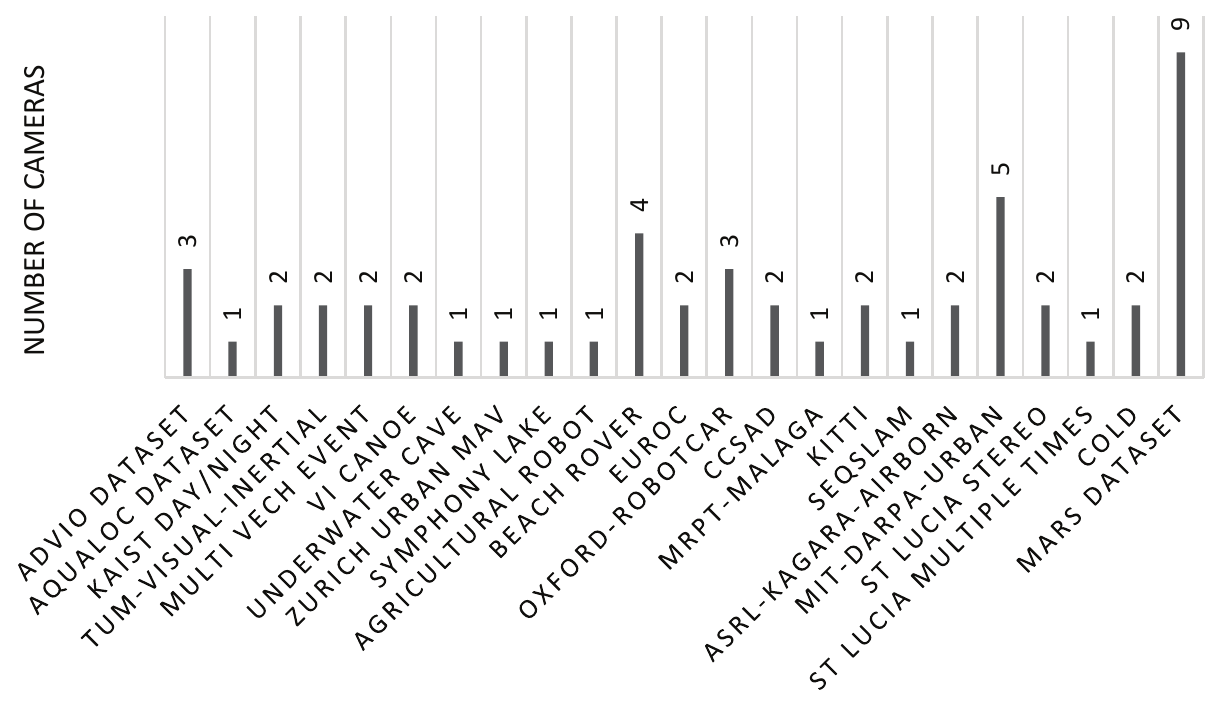}  
    \caption{Number of cameras in SLAM Datasets}
    \label{fig:cameras}
  \end{figure}

\section{Datasets Survey}
 \label{sec:dataset_survey}

 Mapping datasets have a long tradition in the robotics community and there is a big number of datasets for localization and mapping. The datasets can be classified by the environment in which they have been collected: 

  \begin{enumerate}
    \item Indoor: \cite{robothome},\cite{euroc},\cite{TUMRGBD}, provide indoor datasets with annotations.
    \item Outdoor: \cite{KITTI}, \cite{KAISTDayNight},\cite{SPODataset},\cite{ZurichUrbanMAV}, \cite{birk20093d} provide outdoor datasets of outdoor environments.
    \item Simulation: \cite{ICLNUIM} provides a simulation indoor scene dataset. 
  \end{enumerate}

  
The ICL-NUIM dataset \cite{ICLNUIM} mainly focuses on RGB-D cameras with hand-held perspective within simulated indoor environments. The ground truth of this dataset comes from the model of the room. 
  The Robot@Home dataset \cite{robothome} consists of the data stream coming from 4 RGB-D cameras and 2 2D Lidar, but there is no pose ground truth. 
  The TUM-RGBD dataset \cite{TUMRGBD}, as another indoor dataset, is captured by a synchronized Microsoft Kinect RGB-D camera both hand-held and onboard of a ground robot. It has pose ground truth coming from a motion capture system at 100Hz. The images have a resolution of $640\times 480@30$Hz.

  The EuRoC dataset \cite{euroc} is from a micro aerial vehicle (MAV). It is equipped with synchronized stereo monochrome cameras at 20 fps, inertia measurement unit (IMU) and has pose ground truth from laser tracking system.
  The KITTI dataset \cite{KITTI}, as one of the most famous dataset regarding on autonomous driving, is equipped with a HDL-64E 3D Lidar, gray-scale and RGB stereo cameras. All Lidars and the cameras are synchronized and generate data at 10Hz. Global positioning system (GPS) and IMU data are also included in this dataset with 100Hz. The synchronization between GPS and IMU is not hardware-level synchronized with the camera or the Lidar. They use the time stamp to synchronize among them. 
  The Z\"urich Urban MAV dataset \cite{ZurichUrbanMAV} is also based on MAVs. Different from EuRoC, it is recorded in outdoor environments. This capture system has barometer, gyrometer, accelerometer and GPS receiver synchronized with autopilot board and the cameras. Unlike EuRoC, its ground truth comes from algorithms instead of directly from the capture system. GPS location is only used for initial position in its ground truth pose calculation algorithm.

  The CoRBS dataset \cite{CoRBS} is the first to have the ground truth of the camera trajectory and the 3D model of the scenes. The camera is also upgraded to Microsoft Kinect v2 compared to the TUM-RGBD dataset launched in 2012. The RGB resolution is $1920\times 1080@$30Hz and the depth cloud is also collected at 30Hz and havs $512\times 424$ pixels in each frame. The tracking system generates 3D pose ground truth at 120Hz with error of 0.39mm.

  The Oxford RobotCar dataset \cite{Oxfordrobotcar} is a dataset with four types of heterogeneous sensors mounted on a ground vehicle. Its video stream consists of a stereo camera with three lenses, three monocular RGB cameras. It obtains depth data through two 2D Lidars and a four-beam 3D Lidar. The platform is also equipped with a 6 degrees of freedom (DOF) IMU and GPS/GLONASS system reports its position at 50Hz. The three monocular cameras are synchronized at 11.1Hz frame rate, as the stereo camera produces 16Hz. The Lidar scans at 12.5Hz. All these sensors have their clock calibrated using TICSync.

  \cite{birk20093d} provides 3D pointcloud dataset which is recorded in a disater city at 2008 NIST response robot evaluation exercise. Another group of SLAM datasets are simulation datasets.  The noise of the sensors will also affect performance. For simulation datasets it is thus essential to accurately simulate the noise and errors.
 The ICL-NUIM dataset \cite{ICLNUIM} mainly focuses on RGB-D cameras with hand-held perspective within simulated indoor environments. The ground truth of this dataset comes from the model of the room. 
  Datasets of 3D sensor measurements and ground truth poses for benchmarking the performance of SLAM algorithms are provided in \cite{TUMRGBD} and \cite{ICLNUIM}. The ground truth information has been obtained using a tracking system and by creating the data in a simulation, respectively.

  Figure \ref{fig:megapixels} shows a histogram of the total Megapixels in the SLAM datasets of this survey. We can see that most of the datasets provide less than $5$ Megapixels while we provide in total $45$ Megapixels. Figure \ref{fig:cameras} shows the number of cameras in each SLAM dataset. From it we can see that our dataset contains the biggest number of cameras. In Appendix \ref{sec:appendix} we provide a big table listing all mapping datasets we are aware of and a number of their properties, such as number and resolution of cameras, number of laser beams, types of sensors and ground truth information, etc. Attached to this publication is also a file with more details about the datasets.

In Appendix \ref{sec:appendix} we have studied mobile ground mapping robot datasets among all the available datasets, including some recent and named datasets such as Rosario Dataset \cite{RosarioDataset}, Robot@Home \cite{robothome}, and Cheliean Underground \cite{leung2017chilean}. One of the most recent datasets is Rosario Dataset\cite{RosarioDataset}. It is an outdoor mobile ground robot dataset, including dGPS, IMU, stereo cameras, synchronized and has pose ground truth via GPS. Robot@Home \cite{robothome} is an indoor robot dataset, which has a synchronized 2D Lidar stream and relatively low-resolution RGB-D data. Chilean Underground dataset \cite{leung2017chilean} is a slightly different mobile robot dataset. The scenario in it is under the ground. This dataset includes a synchronized stereo camera (two monocular cameras) running at 16Hz with a resolution of 1280*960 and a 3D Lidar. The ground truth pose is from algorithms. The only environmental perception sensor in NCLT dataset \cite{NCLT} is a 32-ring 3D Lidar. As an outdoor dataset, it has GPS, IMU in addition to the Lidar. This dataset is also synchronized and its pose ground truth is obtained through algorithms.

The survey reviled that so far no comprehensive SLAM dataset is available. The requirements are: 
\begin{itemize}
\item High-resolution stereo visual data, preferably in many directions
\item High-resolution 3D Lidar data
\item Fully hardware-synchronized and calibrated
\item Path ground truth information
\item Map ground truth information
\item Diverse environments
\end{itemize}

Thus we created an advanced mapping robot to collect such data. Compared with the aforementioned datasets and referring to Appendix \ref{sec:appendix}, our dataset provides the highest RGB resolution (5 Megapixels for each camera). On the left, front and right side and to the top, we have four pairs of stereo cameras, and a monocular camera looking back. We are also the only dataset to simultaneously have data from stereo cameras, 3D Lidars (one horizontal and one vertical), tracking system (for absolute ground truth poses) as well as 3D ground truth maps (collected from FARO).
Our dataset makes it possible to evaluate both the Lidar and visual odometry against ground truth poses and ground truth maps as well as research on visual odometry and sensor fusion.  

\section{Mapping System}
\label{sec:mapper}

\begin{figure}[t]
	\centering 
	\includegraphics[width=0.8\linewidth]{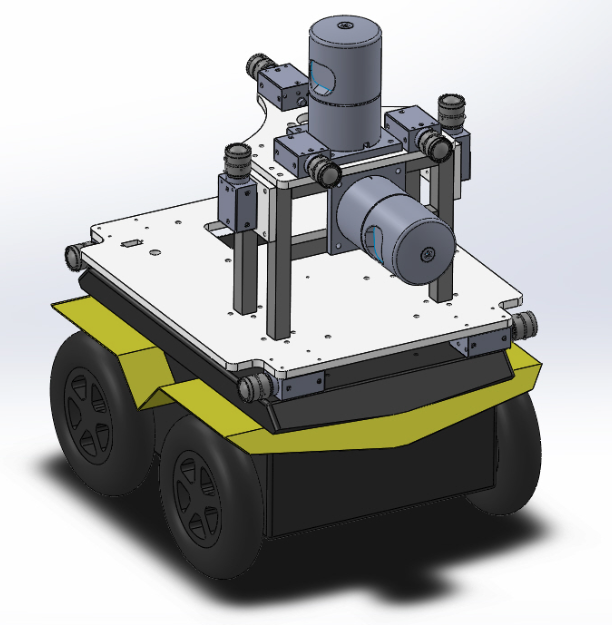}  
	\caption{CAD model of MARS Mapper robot }
	\label{fig:cadrobot_platform}   
\end{figure}

Firstly, a small survey on mapping robots is presented, revealing that our system is one of the most performant mapping dataset collection robots world-wide. Afterwards our robot is described in detail.

\subsection{Mapping Robot Survey}

From the very beginning, mobile robots featured mapping capabilities. As early as 1985 sonar sensors were employed for mapping \cite{moravec1985high} and soon vision and stereo vision \cite{moravec1996robot} systems were used for mapping as well. 

\cite{thrun2000real}, \cite{thrun2003system}  and \cite{mahon2003three} utilize a horizontal 2D Lidar for SLAM and a vertically mounted 2D Lidar for 3D mapping, an approach which is also used in the mapping system presented here, except that we use 3D Lidars with 32 beams each with hardware time-stamping. Another approach is using tilting 2D Lidar sensors for collecting datasets \cite{birk20093}. This has the advantage that it is possible to collect pointclouds with a variable resolution (by tilting slower), but the big disadvantage is, that the robot has to be static while scanning. Continuously rotating 2D lasers are another option for 3D Lidar mapping \cite{digor2010exploration}. \cite{gebre2009remotely} uses a 2D Lidar on a gimbal with a color camera to colorize the 3D pointclouds with no hardware synchronization between the Lidar and the camera. That robot features three wide angle cameras facing forward and backward and one on the gimbal. A mobile mapping robot for underground mines that uses a 64-beam Velodyne which is actively tilted is presented in \cite{neumann2014towards}. In \cite{desai2009objective} different strategies for 3D mapping with 2D Lidars are systematically compared. 

For localization and mapping the fusion of Lidar and camera data can improve the results \cite{iocchi2007building}, \cite{zhang2015visual}. Front and back-facing stereo cameras are also used for visual odometry \cite{oskiper2007visual} and are also combined with a 3D and two 2D Lidars for localization and mapping in \cite{paton2015eyes}. 
Using up-looking cameras for SLAM on the ceiling is also common \cite{jeong2005cv}. Using non-overlapping cameras we can do visual odometry \cite{kneip2013using} and mapping \cite{tribou2015multi}. 


The Oxford RobotCar Dataset \cite{Oxfordrobotcar} was collected using the Oxford RobotCar platform. The RobotCar uses a drive-by-write-capable Nissan LEAF with one Point Grey 
Bumblebee XB3 trinocular stereo camera and three Point Grey Grasshopper2 monocular cameras. 
Additionally, it equips two 2D Lidars and a four-beam 3D Lidar to obtain depth data, a 6 degrees of freedom inertial measurement unit (IMU) and a global positioning system (GPS) 
navigation system.

The Chilean underground mine dataset \cite{leung2017chilean} 
built a platform based on \textit{Clearpath Robotics Husky A200} for collecting the dataset.
The built robot is equipped with a stereo camera, a survey-grade 3D Lidar, and a 
millimeter-wave radar on the upper sensor deck.

The TUM VI benchmark \cite{TUMVisualInertial}, which is a novel dataset with a diverse set of 
sequences in different scenes for evaluating Visual Inertial (VI) odometry, designed system for data collection that contains two cameras in a stereo setup, a microcontroller board with integrated IMU,
a luminance sensor between the cameras and infrared (IR) reflective markers.

The New College Vision and Laser Dataset \cite{smith2009new} designed a vehicle for collecting
data. It consists of two lasers scanners in the body-vertical plane on the sides of the vehicle, a
LadyBug Panoramic camera for collecting five-view omnidirectional imagery and a BumbleBee stereo camera pair.

The KAIST Multi-Spectral Day/Night Data Set \cite{KAISTDayNight} developed a multi-sensor platform gathering data for autonomous and assisted driving. The capturing system is mounted on top of a 
vehicle, which equipped with two PointGrey Flea3 RGB cameras, a FLIR A655Sc thermal camera,
a Velodyne HDL-32E 3D LiDAR and a GPS navigation system.

In addition to the mapping robots whose working environment is ground, there are plenty of mapping robots
working in the air or water.

The EuRoC MAV Datasets \cite{euroc} was collected by a Micro Aerial Vehicle (MAV). It is an AscTec Firefly 
MAV2 equipped with visual-inertial sensor unit in a front-down looking position, two global-shutter and  
monochrome cameras as well as a time-synchronized IMU.

The Visual–Inertial Canoe Dataset \cite{VICanoe} gathered data by using a canoe. The canoe was equipped with a stereo camera, an IMU, and a GPS device, which provide visual data suitable for stereo or monocular applications, inertial measurements, and position data for ground truth. 

None of the mapping robots we presented here or found in literature offers as high-resolution image data as our robot, and we are on par with the best robot w.r.t. the Lidar sensors and IMU. In so far we see our claim justified to have build one of the most performant mapping robots worldwide. 

\subsection{Mapping Robot Description}

We are presenting a fully hardware synchronized mapping robot with support for a hardware synchronized external tracking system, for super-precise timing and localization. The vehicle is equipped with nine high-resolution cameras and two 32-beam 3D Lidars based on Jackal Robot. A professional, static 3D scanner is used for ground truth map collection.

The advanced mapping robot used to collect the datasets for this paper is designed to collect as much data as possible for indoor mapping scenarios. We collect data from vertical and horizontal Lidars, stereo cameras, an IMU and the robot odometry. The MARS mapping robot's base is a Clearpath Jackal differential drive robot with an upgraded power supply and computer (Intel Core i7-6770k CPU, Raid 0: 3x Samsung 850 EVO 500G). The mainboard supports eight independent USB 3.1 ports, which are mainly used for the cameras. The robot is collecting data from the following sensors:

\begin{itemize}
	\item Nine 5MP wide-angle color cameras (FLIR Grasshopper3 GS3-U3-51S5C-C) with wide-angle lenses (82$^{\circ}$ x 61$^{\circ}$), 10Hz (4 stereo pairs: front, left, right, up; one back-looking camera)
	\item Two Velodyne HDL-32E 3D Laser scanners, 10Hz (one horizontal, one vertical; both in dual-return mode)
	\item IMU (Inertial Measurement Unit): Xsense MTi-300, 200Hz
	\item Robot odometry
	\item Optitrack tracking system (21 Prime 13 cameras, 30Hz)
\end{itemize}

A CAD model of our robot showing the sensors is depicted in Figure \ref{fig:cadrobot_platform}. We compress the camera images with JPEG quality 90. Due to CPU speed limitations, we can not store much more than 10Hz for the 9 cameras, so we chose to collect the images with the same frequency as the Velodynes. This results in a total storage bandwidth of about 170 MB/s.

\section{Sensor Synchronization}
\label{sec:sync}

 Almost all famous datasets are collected with platforms that contain several types of sensors. Most of the datasets contain cameras, Lidars, and IMU/GPS information. Data synchronization is quite important for many robot tasks such as localization and navigation. To achieve precise localization and safe navigation, several types of sensor data are usually needed. These sensors can be divided into two groups, the first group is sensors can be triggered externally or can provide an individual counter by each measurement. The second category are sensors that can not be externally triggered, such as Lidars.  
 \cite{wu2018soft} proposes a software time synchronization framework to synchronize all sensors. They use ROS to synchronize all sensors, but their method is limited by the minimum frequency of the sensor.

 \cite{hwang2015low} propose a low cost method to synchronize multi-cameras. One camera is used as master to trigger other slaves cameras. \cite{olson2010passive} provide a passive synchronization solution to eliminate the data translation error. \cite{litos2006synchronous} present a software time synchronization method for multi-cameras. The client-server method is utilized for real-time synchronization of multi-cameras. Two computers are connected by  Ethernet and the time of them are synchronized by NTP\cite{mills1991internet}. Finally, an LED array system is used to verify the effectiveness of this approach. Similar to \cite{litos2006synchronous}, we use an extra SoC (System on a Chip) as an independent clock source to generate phase-aligned square waves as clock pulses to trigger sensors.

 \begin{figure}[t]
	\centering 
	\includegraphics[width=0.8\linewidth]{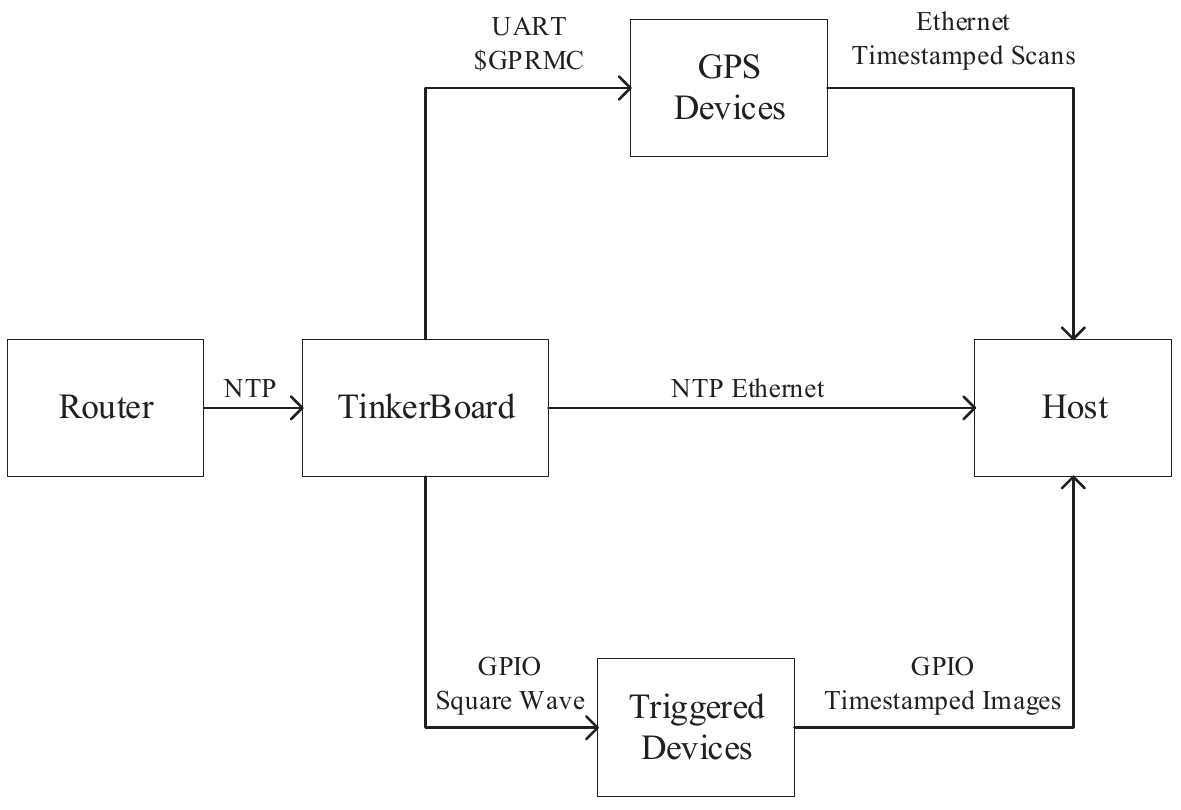}  
	\caption{Trigger \& Timestamp Spread Topology }
	\label{fig:Topology}
\end{figure}

\begin{figure}[t]
	\centering 
	
	\includegraphics[width=0.4\linewidth]{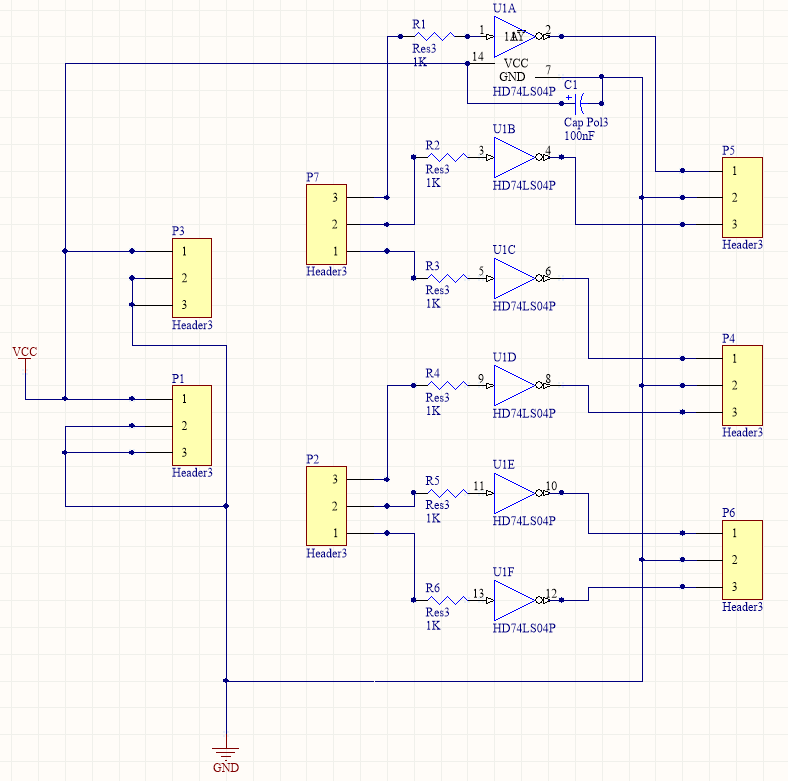}  
	\includegraphics[width=0.3\linewidth]{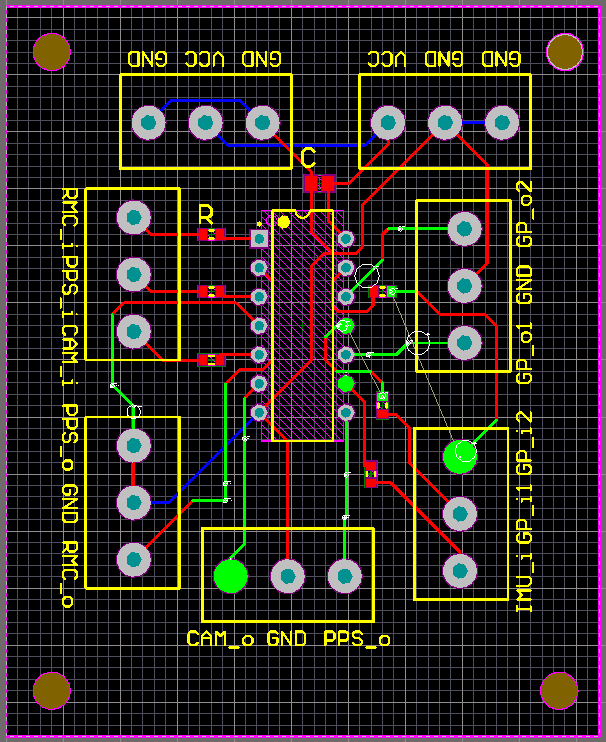}  
	\caption{PCB schematic diagram of our synchronization board }
	\label{fig:pcb}
\end{figure}

We are interested in collecting data with very high accuracy for very precise mapping, so it is essential to synchronize all the sensor data by hardware. We use an Asus Tinker Board with a quad-core 1.8 GHz ARM Cortex-A17 processor to provide hardware synchronization. The Tinker Board serves as our reference time. It is triggering the cameras and the tracking system with 10 Hz and the Xsens IMU with 200Hz. But hardware triggering of sensors is only half of the job: Afterwards the data from the cameras arrives on PC in the Jackal robot at different times, due to USB and CPU scheduling issues. We thus need to be able to associate the hardware triggers with the actual data and make use of this info in the software. Thus, for every trigger it generates, the Tinker board also sends a ROS timestamp to the PC, which is collected in the bagfile. In a post-processing step, we then match the sensor data with this time-stamp and then correct the time-stamps of the data. Afterward, all data (e.g. all images and the tracking data and the IMU information) that was triggered together will have exactly the same time-stamp.

 The tracking system is also triggered by the Tinker board, but it is then increasing the frequency from 10Hz to 30Hz. For that, the robot is physically connected to the tracking system via an Ethernet cable when inside the systems camera view. Before leaving or entering the tracking system we manually (un)plugged this cable, briefly stopping the data-collection for this. To avoid having a second Ethernet cable from the tracking system to the robot, we collect the tracking system's (which is running on Windows) data on a separate Ubuntu PC. Before each run, the time of the PC, the Jackal PC and the Tinker Board are updated via NTP from the router of our lab. 

Since the Velodyne is a rotating sensor, it cannot be triggered. Instead, its time-stamps are messages using GPS (Global Positioning System) pps (pulse per second) and NMEA (National Marine Electronics Association) data which is encoding the time. The Tinker Board is providing fake GPS data with its own time to the Velodynes, such that their data arrives at the Jackal PC already with the correct time stamp. 

To achieve synchronization among several heterogeneous sensors and the host for the data frames, we require extra circuits and timing topology for hardware-level synchronization and some auxiliary algorithms for software-level synchronization. The hardware-level synchronization is shown in Figure \ref{fig:Topology}.


%
%
%
%
%

\subsection{ Hardware Timing Topology}

  We classified the devices to be synchronized into two classes: triggered devices and GPS-stamp devices. For the triggered devices (e.g. cameras and IMUs), we generate square waves (whose initial phases are aligned to second borders) and deliver the waves to the devices as triggering signal. For the GPS-stamp devices (e.g. Velodyne Lidars), we use a 1Hz square waves with their duty cycle adapted to specified targeting devices as PPS via GPIO (General Purpose Input Output) and timestamps encoded to GPRMC (a NMEA 2.0 sentence) format via UART (Universal Asynchronous Receiver/ Transmitter). The delays in all of these signal transmissions are measured to be less than 20us.
  
We designed a simple configurable voltage conversion circuit which can convert 3.5V to 5V as shown in Figure \ref{fig:pcb}. It is able to hardware synchronize multiple cameras and multiple Lidars, which requires different signal patterns simultaneously at low cost and affordable complexity without extra facilities.
The whole voltage conversion circuit is assembled on a printed circuit board (PCB).

%
%
%
%

\subsection{ Tests and Verifications}

Table \ref{table:my-table} shows the time delay between the different channels of signal. For example, $"$10Hz $\&$ 30Hz$"$ means the time delay between 10Hz signal and 30Hz signal. 20 groups were tested for each set of signals, and the mean and variance were calculated. Limited by the measuring instrument, the measurement accuracy is 400ns.

\begin{table}[]
\caption{Time delay between different signals}
\label{table:my-table}
\begin{tabular}{|l|l|l|l|}
\hline
           & 10Hz \& 30Hz & 10Hz \& PPS & 30Hz \& PPS \\ \hline
average(ns) & 15120        & 0           & 3720        \\ \hline
variance   & 1.36E+09     & 0           & 2.77E+08    \\ \hline
\end{tabular}
\end{table}

From the data in Table \ref{table:my-table} we can see that the error is controlled within 20 microseconds, which is far smaller than 30 milliseconds for 30Hz. So, this satisfies our need for hardware sensor synchronization.



\section{Sensor Calibration}
\label{sec:calibration}

\begin{figure}[tb]
		\centering
		\includegraphics[width=1.0\linewidth]{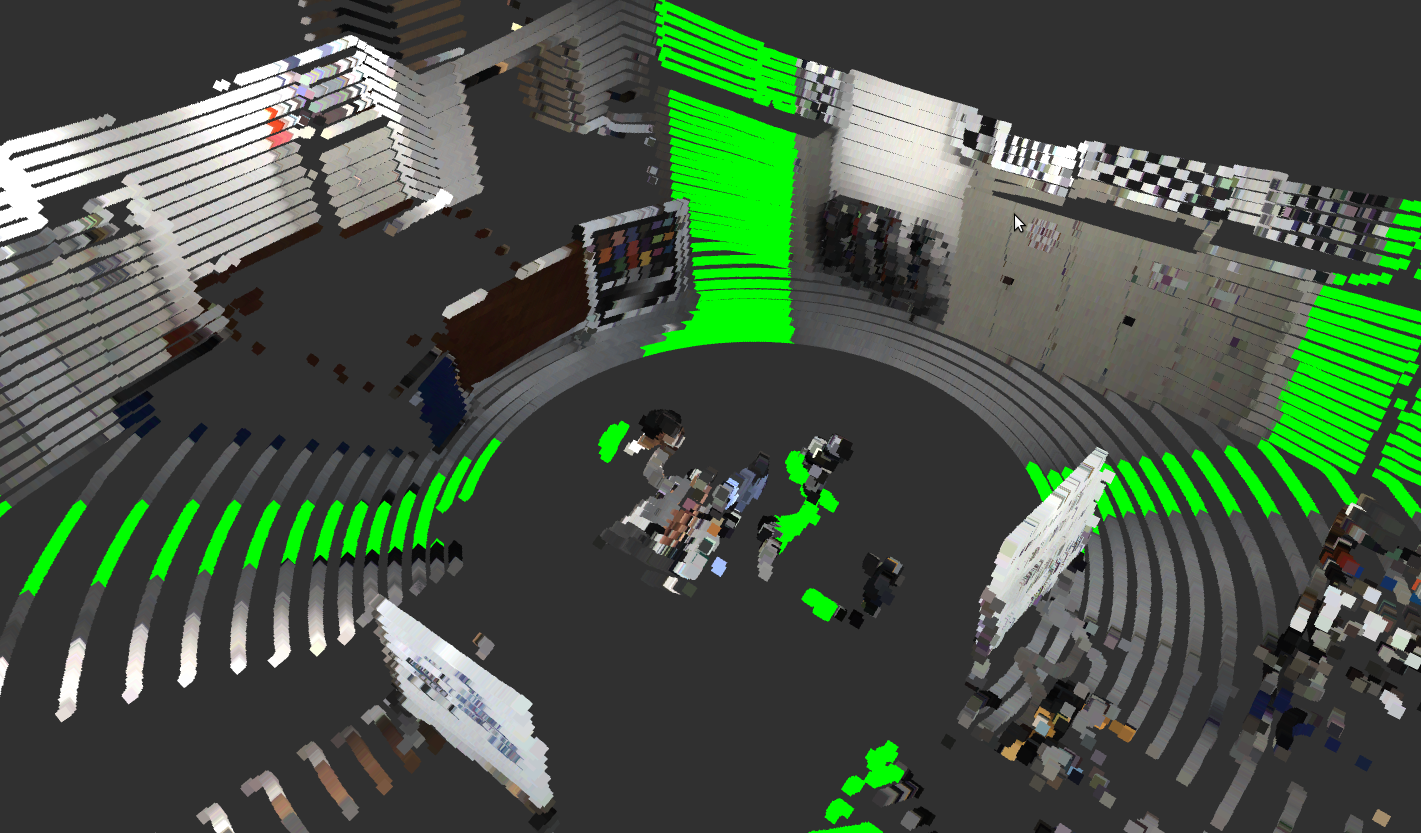}  
	\caption{ A 3D Lidar scan colored by all the $7$ horizontal cameras. The transformations between each camera and 3D Lidar are acquired from global optimization result. All the green points represent areas where no camera is overlapping with the pointcloud. }
   \label{fig:coloredpoints}
\end{figure}

\begin{figure}
	\begin{minipage}[t]{0.5\linewidth}
		\centering
		\includegraphics[width=1.6in]{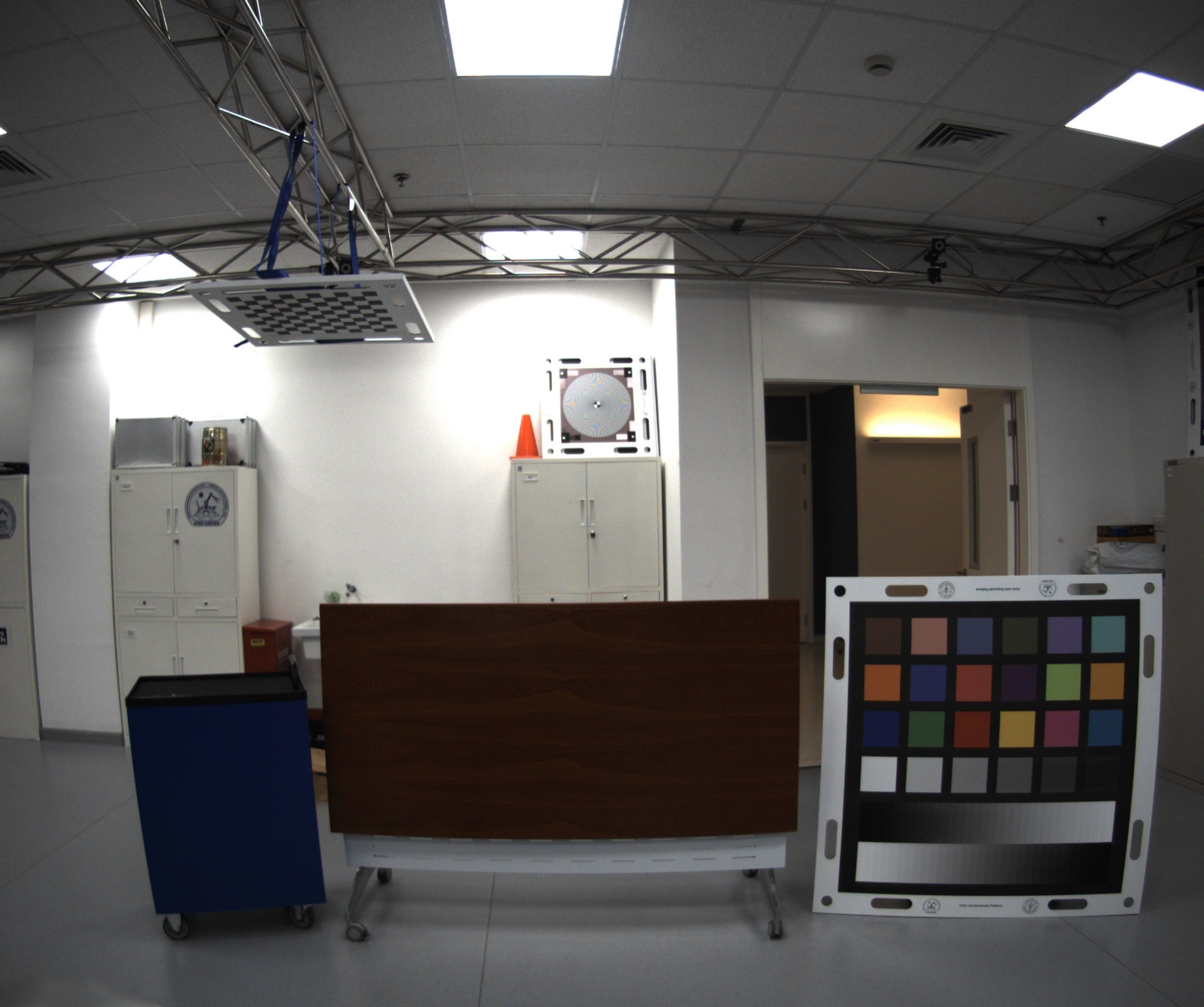}
	\end{minipage}%
	\begin{minipage}[t]{0.5\linewidth}
		\centering
		\includegraphics[width=1.6in]{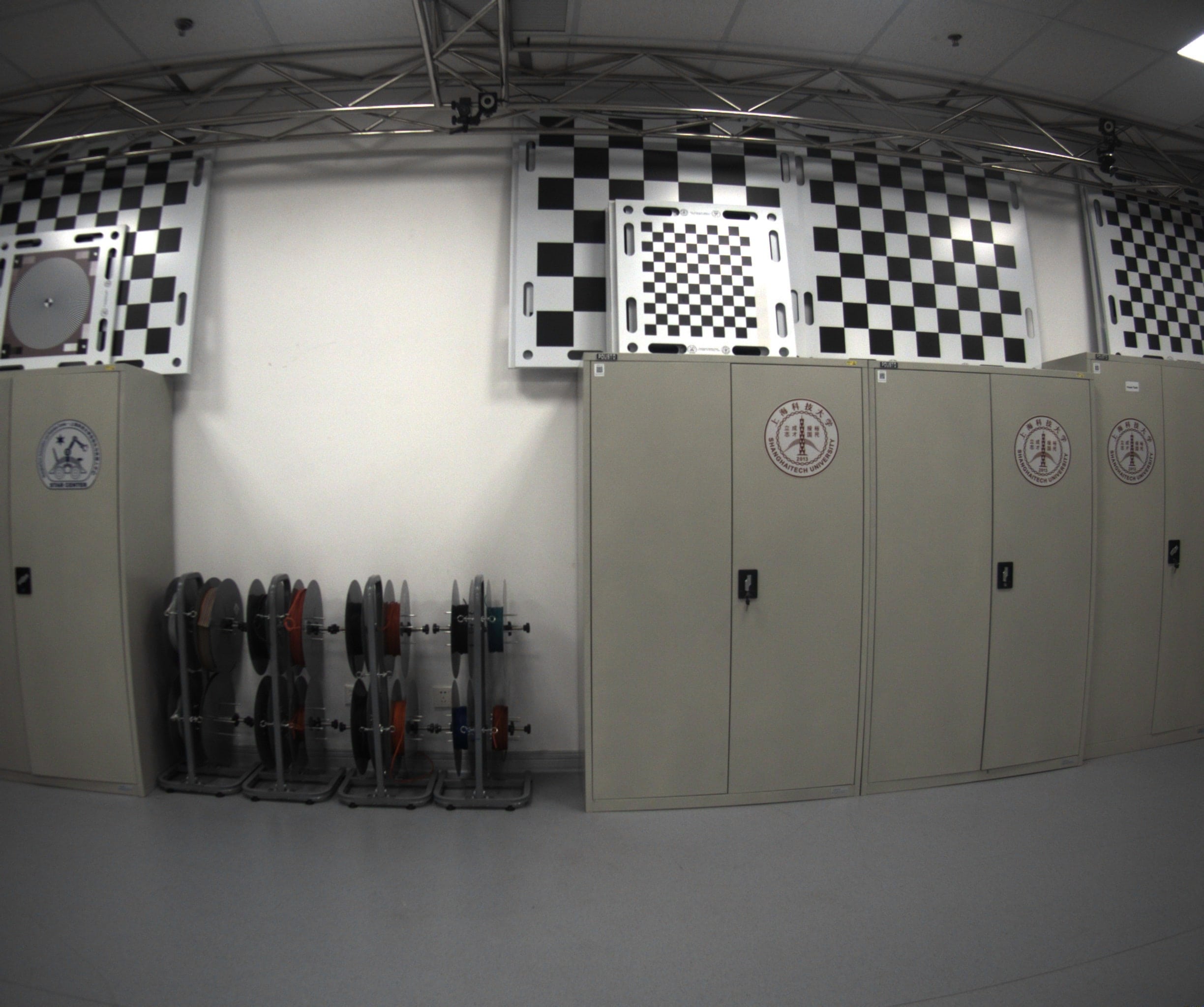}
		
	\end{minipage}
	\caption{Images acquired from two cameras. The left image is acquired from one of the left side cameras while the  right image is acquired from one of the front cameras.}
	\label{fig:normalimage}
\end{figure}

As outlined above, mapping robots are typically equipped with different kinds of sensors. Some complex tasks need to fuse the data from the different sensors. For instance, cameras can provide rich RGB information, while odometry provides the trajectory of the robot. In order to fuse these data, the extrinsic parameters of each sensor must be calibrated.  Calibration of sensors has been studied for a long time. Generally, calibration methods can be divided into two groups.
 
 The first group is single pair sensor calibration. Single pair calibration means to compute the transformation between two sensors. \cite{Dhall2017LiDAR} proposes an approach to compute the relative pose of one camera with 3D Lidar by using 3D-3D correspondence. \cite{daniilidis1999handeye} compute extrinsic parameters of one camera and robot arm by using dual quaternions. Our group presented work on simultaneous hand-eye calibration and reconstruction in \cite{Zhi2017Simultaneous}.  \cite{Zhang2005Extrinsic} proposes a method to calibrate one 2D Lidar and one camera. \cite{Zhang2000A} provides an easy way to calibrate cameras. 
 
 \begin{figure}[b]
\centering 
\includegraphics[width=0.5\linewidth]{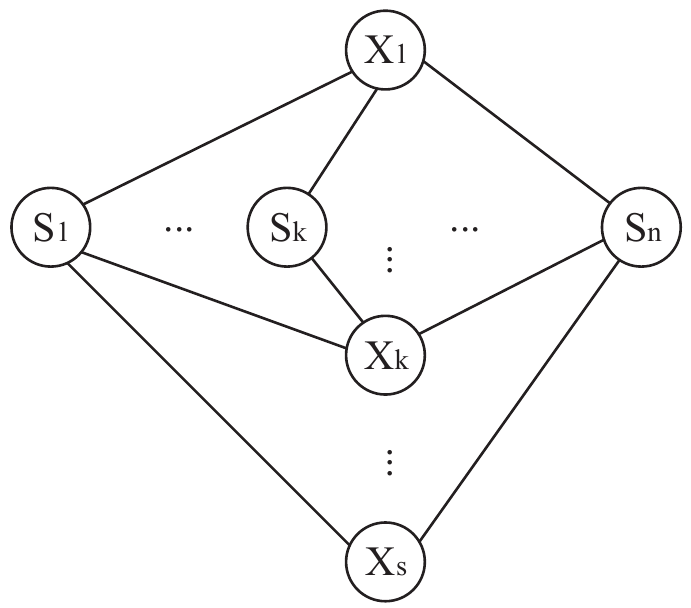}  
\caption{
	Node $X_1 ...  X_s$ and node $ S_1 ...  S_n $ are different types of sensors such as camera and Velodyne.  An edge between two nodes represents a direct sensor-to-sensor calibration between these two devices.}
\label{fig:posge_graph} 
\end{figure}

 Another group is multi-sensor calibration. As the number of sensors on the robot increases, the importance of multi-sensor calibration becomes higher and higher. \cite{heng2013camodocal} propose a method to calibrate the extrinsic parameters of multi-cameras by using a bundle adjustment approach. \cite{sim2014closed} uses a closed loop constraint to calibrate multi-sensors. \cite{chy230} uses a graph-based method to calibrate the relative pose of each sensor. First, extrinsic parameters between each sensor pair are calculated, after that, a graph-based method is used to reduce the global error. We adopt their approach to calibrate all the sensors mounted on our robot.

An essential assumption for most calibration approaches is, that all sensors are rigidly mounted on the robot platform. Extrinsic calibration then means the estimation of the relative poses of sensor pairs, such that all the data collected from the different sensors can be fused into one single frame.  Due to sensor noise, it is impossible to align the data without error. Also, when using real sensors, there is typically no way to accurately measure the transform (translation and rotation) of the physical sensor to another frame. This is because the sensor frame is usually somewhere inside the sensor, and because there are no tools available to measure arbitrary translations and rotations of physical objects with sufficient accuracy.

 Our MARS Mapper robot is fully calibrated. For sensor calibration, we use the algorithm which is proposed in \cite{chy230}. Intrinsic calibrations are acquired using known methods. The extrinsic calibration of the sensors (i.e. their poses) are gathered by pair-wise calibration of various, also heterogeneous sensor pairs (4x stereo cameras, 32x non-overlapping cameras, 13x Lidar to the camera, 1x Lidar to Lidar, 9x tracking system to camera) and then minimizing the error using G2O \cite{kummerle2011g}.
Once relative poses between each sensor are known, by using \cite{chy230}'s algorithm, a hypergraph is build to minimize the global calibration error.
Figure \ref{fig:posge_graph} shows the relationship between  different sensors. As the algorithm describes in \cite{chy230}, a global error function is defined in Eq. \ref{globalerror}.

\begin{equation}
F(x) = \sum_{v_i,v_j\in V} e(x_i, x_j, u_{ij})^T \Omega_{ij} e(x_i, x_j, u_{ij})
\label{globalerror}
\end{equation}

\begin{equation}
\label{formula}
\hat{x} = \mathop {\arg\min }\limits_x  F(x)
\end{equation}

where  $u_{ij}$ is the initial constraint of node $i$ and $j$. $\Omega_{ij}$ represents the information matrix of the constraint.

With an initial guess $\hat{x}$, the solution of Eq. \ref{formula} can be found by incrementally by solving a linear system with the system matrix $H$ and the vector $b$, such that 
\begin{equation}
H = \sum_{i, j \in V} J_{ij(\hat{x})}^T \Omega_{ij} J_{ij(\hat{x})}
\end{equation}
\begin{equation}
b^T = \sum_{i, j \in V} e_{ij}^T \Omega_{ij} J_{ij(\hat{x})}
\end{equation}. 
Here, $J$ is the Jacobian of the error function, with an initial guess $\hat{x}$. To effectively solve the non-linear function we use G2O \cite{kummerle2011g}.

\begin{figure*}[t]
	\centering
	\includegraphics[width=0.84\linewidth]{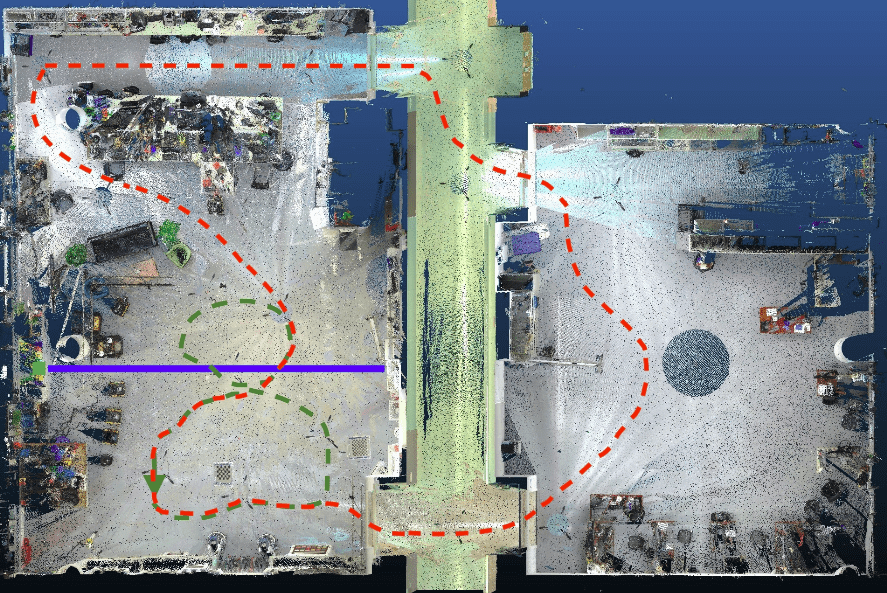}  
	\caption{Top-view of the ground-truth faro pointcloud (480mill points; ceiling points were removed to provide view inside the rooms) with the MARS-8 path (green), the MARS-Loop and MARS-NoLoop paths (red) 
		as well as the location of the curtain (blue) added.}
	\label{fig:paths}
\end{figure*}

\begin{figure*}[h!]
	\centering
	\includegraphics[width=0.43\linewidth]{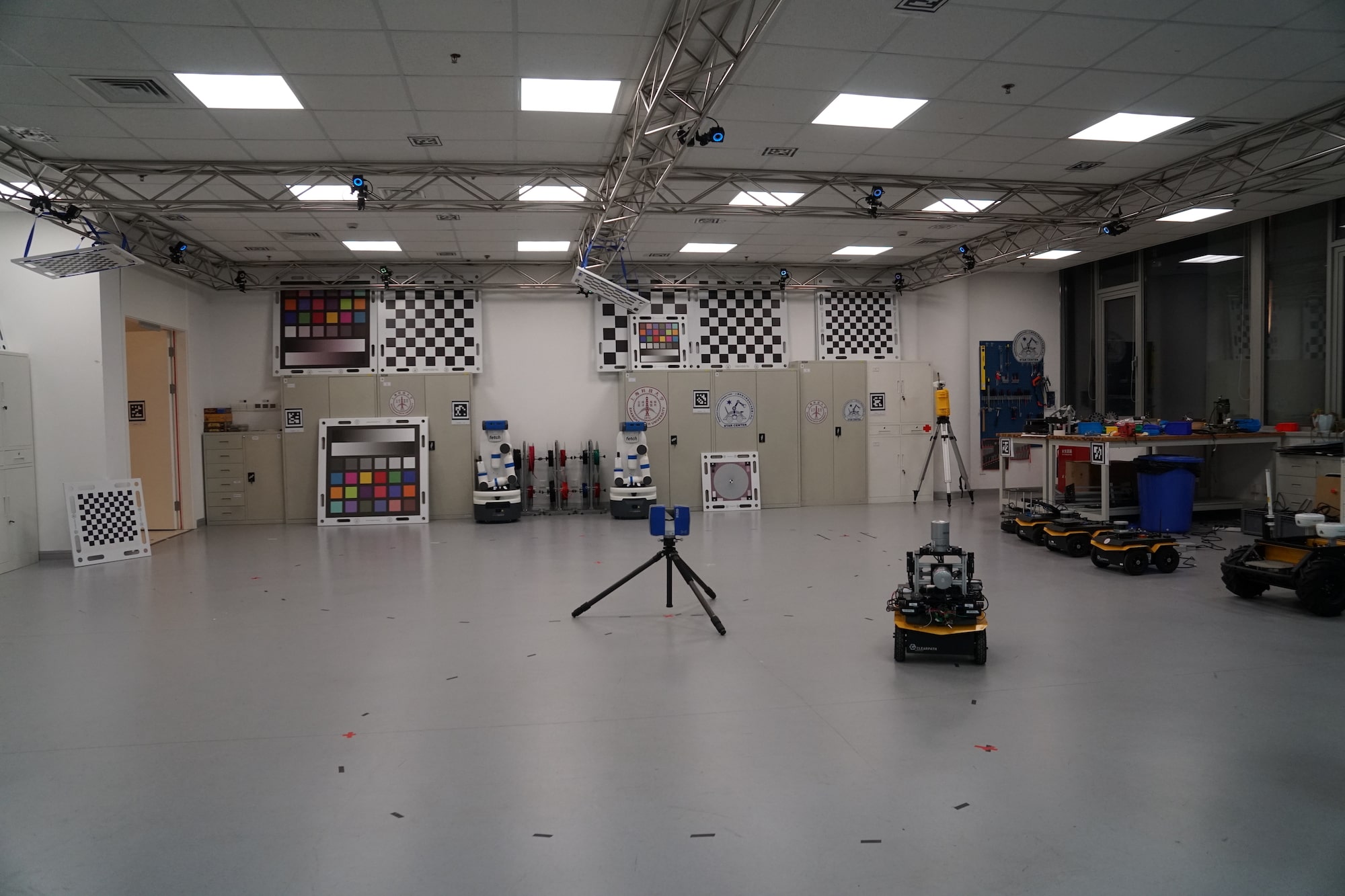}   \vspace{2mm}
	\includegraphics[width=0.43\linewidth]{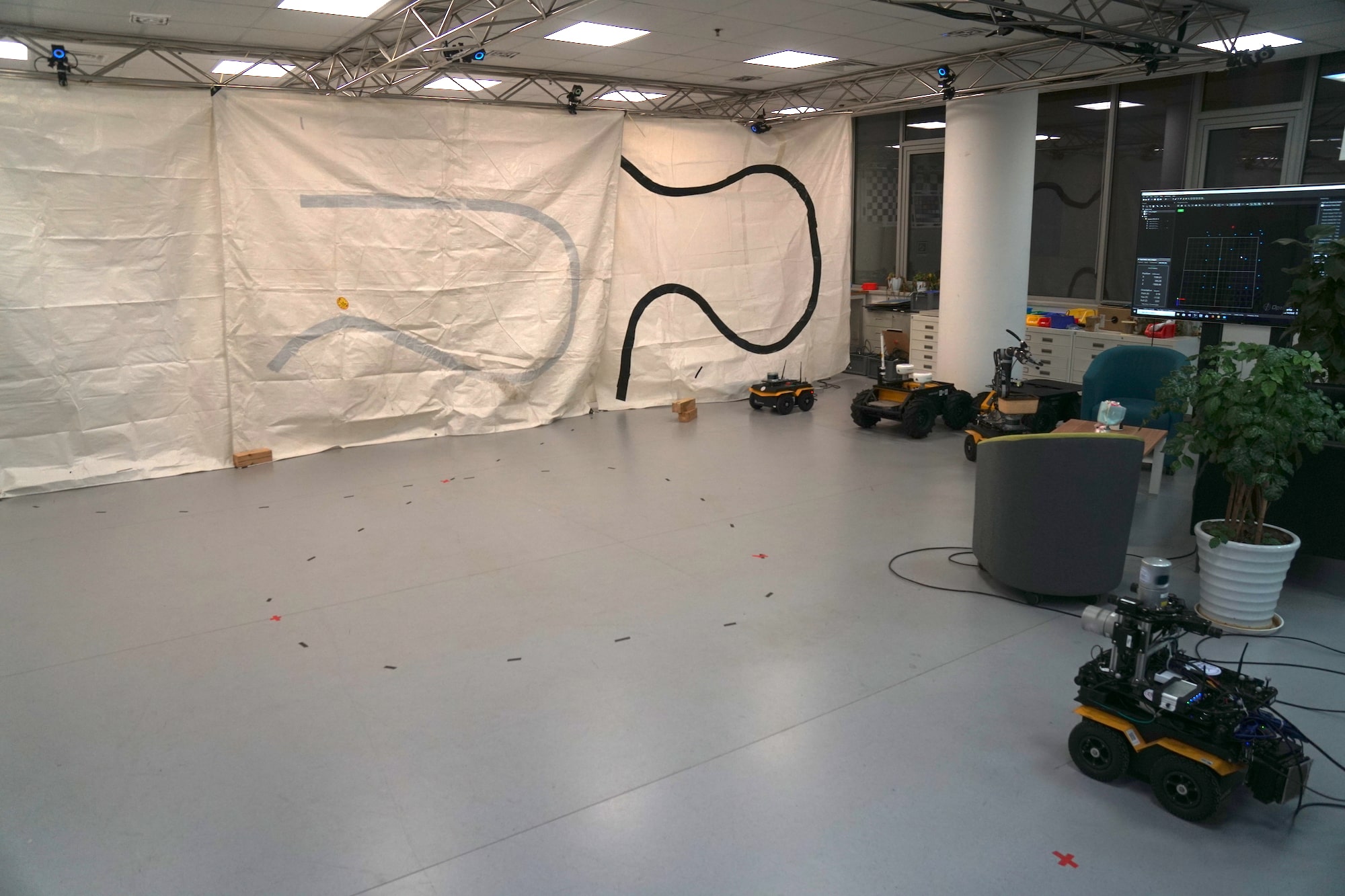}  
	\caption{The robot and the FARO scanner in the MARS lab on the left. On the right: the same area with the curtain (made with tarp) for map MARS-NoLoop.   }
	\label{fig:lab}
\end{figure*}

An application of sensor fusion that requires accurate multi-sensor calibration is the colorization of Lidar point clouds using the cameras. Figure \ref{fig:coloredpoints} shows a Velodyne scan from our robot, where the points that are within the field of view of one of the 7 horizontal cameras are colored, while Figure \ref{fig:normalimage} shows two of the images used for the colorization. The according code can be found on the dataset webpage in the BLAM package. In Figure \ref{fig:colorBLAM} this data is used to build 3D maps of colored points.


\section{Datasets}
\label{sec:datasets}

\begin{table}[]
    \caption{Dataset Details}

    \centering
    \begin{tabular}{p{4cm}p{1cm}p{1cm}}
        \hline
        Name  & Duration(s) & Size(GB) \\
        \hline
        MARS-8 & 99   & 16.4                                      \\
        MARS-Loop  &290    & 50.7                                     \\
        MARS-NoLoop  & 315   & 54.8                                     \\
        MARS-8-Sample &3 & 0.49\\
        
        \hline       
        3D FARO raw ground truth & &2.3 \\
        3D FARO subsampled & &0.2\\
        2D ground truth map & &$<$0.001\\
        ground truth trajectories & &$<$0.001\\

    \end{tabular}
    \label{bs2}
\end{table}

\begin{figure}[tb]
	\centering
	
	\includegraphics[width=0.7\linewidth]{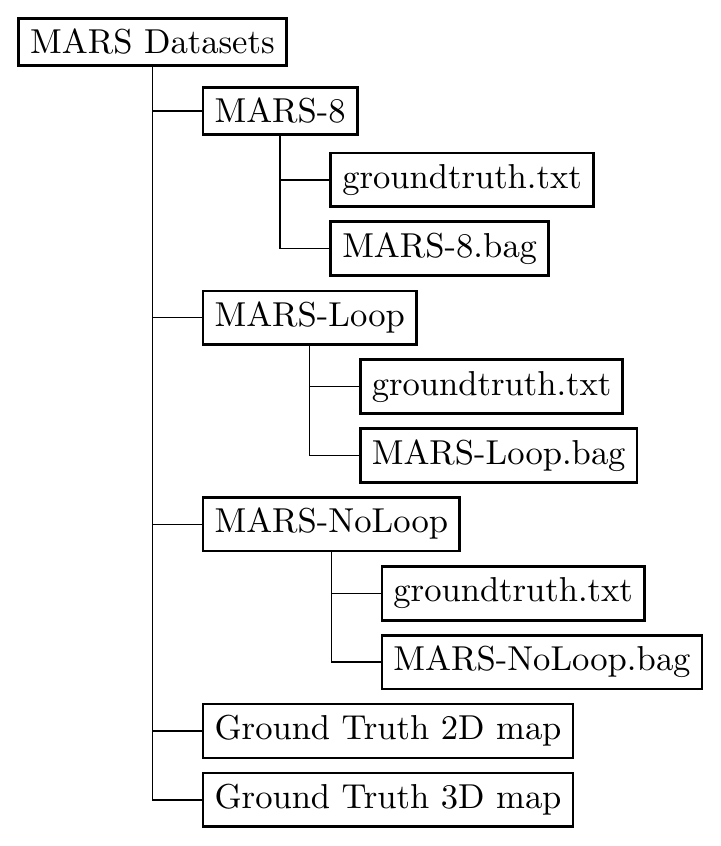}  

	\caption{The data format of our dataset}
	\label{fig:bagtree}
	
\end{figure}

        \addtolength\abovecaptionskip{-9pt}

\begin{figure*}[tb]
	\centering

	\footnotesize	
	\begin{verbatim}
                Topic Name                 Number of Messages       Topic Type
    /camera_back/camera_info                     996 msgs    : sensor_msgs/CameraInfo      
    /camera_back/image_raw/compressed            996 msgs    : sensor_msgs/CompressedImage 
    /camera_front_l/camera_info                  995 msgs    : sensor_msgs/CameraInfo      
    /camera_front_l/image_raw/compressed         995 msgs    : sensor_msgs/CompressedImage 
    /camera_front_r/camera_info                  995 msgs    : sensor_msgs/CameraInfo      
    /camera_front_r/image_raw/compressed         995 msgs    : sensor_msgs/CompressedImage 
    /camera_left_back/camera_info                995 msgs    : sensor_msgs/CameraInfo      
    /camera_left_back/image_raw/compressed       995 msgs    : sensor_msgs/CompressedImage 
    /camera_left_front/camera_info               995 msgs    : sensor_msgs/CameraInfo      
    /camera_left_front/image_raw/compressed      995 msgs    : sensor_msgs/CompressedImage 
    /camera_right_back/camera_info               995 msgs    : sensor_msgs/CameraInfo      
    /camera_right_back/image_raw/compressed      995 msgs    : sensor_msgs/CompressedImage 
    /camera_right_front/camera_info              995 msgs    : sensor_msgs/CameraInfo      
    /camera_right_front/image_raw/compressed     995 msgs    : sensor_msgs/CompressedImage 
    /camera_sync                                 996 msgs    : std_msgs/Time               
    /camera_up_l/camera_info                     995 msgs    : sensor_msgs/CameraInfo      
    /camera_up_l/image_raw/compressed            995 msgs    : sensor_msgs/CompressedImage 
    /camera_up_r/camera_info                     995 msgs    : sensor_msgs/CameraInfo      
    /camera_up_r/image_raw/compressed            995 msgs    : sensor_msgs/CompressedImage 
    /horizontal_velodyne/velodyne_packets        995 msgs    : velodyne_msgs/VelodyneScan  
    /horizontal_velodyne/velodyne_points         994 msgs    : sensor_msgs/PointCloud2     
    /imu/data                                   7289 msgs    : sensor_msgs/Imu             
    /imu/data_raw                               7289 msgs    : sensor_msgs/Imu             
    /imu/mag                                    7289 msgs    : geometry_msgs/Vector3Stamped
    /imu_sync                                  19913 msgs    : std_msgs/Time               
    /jackal_velocity_controller/odom            4978 msgs    : nav_msgs/Odometry           
    /odometry/filtered                          4978 msgs    : nav_msgs/Odometry           
    /tf                                         8154 msgs    : tf2_msgs/TFMessage          
    /tf_static                                     1 msg     : tf2_msgs/TFMessage          
    /vertical_velodyne/velodyne_packets          995 msgs    : velodyne_msgs/VelodyneScan  
    /vertical_velodyne/velodyne_points           994 msgs    : sensor_msgs/PointCloud2     
    /xsens_imu/data                            19912 msgs    : sensor_msgs/Imu             
    /xsens_imu/velocity                        19912 msgs    : geometry_msgs/TwistStamped
             \end{verbatim}
	
	\caption{Data contained in the rosbag of the MARS-8 dataset.}
	\label{fig:baginfo}
	
\end{figure*}

\begin{figure*}[tb]
	\centering

	\tiny  
    \begin{verbatim}
                    # timestamp tx ty tz qx qy qz qw
                    1552558895.519010066986 -0.043429993093 0.065021090209 0.327500402927 -0.001032393775 -0.001503689447 -0.129771366715 0.991542279720
                    1552558895.586153984070 -0.043425511569 0.065015487373 0.327505290508 -0.001031934284 -0.001496045152 -0.129786849022 0.991540312767
                    1552558895.586299657822 -0.043406683952 0.065024986863 0.327479898930 -0.000972410548 -0.001309979358 -0.129725992680 0.991548538208
                    1552558895.618683338165 -0.043386425823 0.065008759499 0.327457219362 -0.000923991203 -0.001135770348 -0.129697501659 0.991552591324
     \end{verbatim}
	
	\caption{The ASCII ground truth trajectory data format.}
	\label{fig:traj}
	
\end{figure*}

        \addtolength\abovecaptionskip{9pt}

Using the synchronized and calibrated MARS mapping robot we collected three datasets in the Mobile Autonomous Robotic Systems Lab (MARS Lab) of ShanghaiTech University:

\begin{itemize}
  \item \textbf{MARS-8}: A short (23m) figure eight driven by the mapping robot with continuous tracking information. For basic SLAM evaluation and evaluation of mapping performance.
  \item \textbf{MARS-Loop}: A medium length (77m) mapping run, starting in the tracking system in the MARS lab, leaving the lab and re-entering it through a different door, finally entering the tracking system again and finishing at the start pose. For evaluation of basic loop closing performance.
  \item \textbf{MARS-NoLoop}: The MARS lab is divided into two parts by two curtains (10cm apart; along the center of the tracking system). The robot follows the same path as MARS-Loop, except that it stops a little earlier (because the curtains are in the way). The robot starts and ends in the same tracking system. No loop closing is possible between the start and end of the dataset, because there is almost no overlap between the areas. 
\end{itemize}

The datasets were collected by driving the robots teleoperated, using WiFi to transmit the front camera and joystick data via ROS messages.

Figure \ref{fig:paths} shows the paths of the robot in the different datasets: green for MARS-8 and red for MARS-Loop and MARS-NoLoop. The approximate paths we followed are also marked on the ground and are thus visible in some of the collected camera images (black for MARS-8 (and MARS-Loop where they overlap) and white for MARS-Loop). MARS-NoLoop is following MARS-Loop, except stopping earlier. Almost nothing in the environment was changed between the robot data collections nor for the Faro scans.

Figure \ref{fig:paths} is actually the complete pointcloud (480 million points) from the 18 FARO scans we collected (each about 27 million points). The points from the ceiling were removed in order to provide the view inside the rooms. We used the FARO Scene software to register the scans. It reported an average error of the scan points of 1.2mm. This is an excellent value and much smaller than the expected sensor noise. The FARO data can thus serve as ground truth for map comparison. The approximate positions of the Faro scans are marked with red crosses on the ground. Most of the scans were taken at a hight similar to the horizontal Velodyne (61 cm). Figure \ref{fig:lab} shows the MARS Lab with the markings on the floor, the Faro scanner and the robot. It also shows the curtain for MARS-NoLoop.

We also placed several checkerboards in the lab. Additionally, we have many April tags distributed on the ceiling and, in the MARS Lab, also on the walls. In the future, we plan to evaluate how well those can be used for localization evaluation of SLAM algorithms. For good measure we also placed other cool robots of the MARS Lab as well as a small living-room arrangement with sofa, plants and TV in the scene. The truss and cameras of the tracking system can also be seen in Figure \ref{fig:lab}.
Table \ref{bs2} shows details about our datasets.

\subsection{High-resolution Data}

From Table \ref{bs2} we observe that the size of the dataset is very big - in total over 120GB. This amount of data is a challenge to store, transmit and process. In so far, we are now justifying working with such big data:

\begin{itemize}
  \item Future robots will be able to afford high-resolution sensors and support very high computation capabilities, we thus want to provide already now the data that future robots might work with, in order to already develop the according algorithms. 
  \item One aim of this dataset is to answer scientifically if using high resolution data (e.g. 5MP videos) is overkill, and if, how much so. Specifically, the data should enable researchers to answer the question: What are the benefits of high resolution sensors, compared to their cost (computation and memory). We can only do this by having that high resolution data and doing and evaluating the according experiments.
  \item In order to answer the above question, if high resolution data is useful, comparison with lower resolution data is needed. We argue that the best way of obtaining the lower resolution data is by sub-sampling the high resolution data (w.r.t. image size, Lidar point number, frame rate), for several reasons:
  \begin{itemize} 
  \item The difference between sub-sampled high-resolution data and low-resolution data from low-resolution sensors (or high-resolution sensors with a low-resolution setting) is quite small. 
  \item In contrast, collecting another dataset during a different mapping run may result in slightly different trajectories, different exposure, focus or white-balancing settings or even changes in the environment. Those could potentially have big impact on the result of the mapping run. So we very much prefer sub-sampling over taking another, dedicated low-resolution dataset.
  \item Sub-sampling allows to test many different resolutions, while restricting the experiments to the resolution they were actually taken in, only offers those very few resolutions.
  \item It is much cheaper to collect the dataset just once.
  \end{itemize}
\end{itemize}

\subsection{Dataset Format}
The three datasets are collected with our mapping robot, using the Robot Operation System (ROS). All datasets are provided as ROS bag files. Figure \ref{fig:bagtree} shows the format of our MARS dataset in terms of ROS topics and message types, on the example of MARS-8. For each dataset, the file {\it groundtruth.txt} contains the ground truth trajectory information.  All message types (data types) in the bag files are standard ROS message. Figure \ref{fig:traj} shows the data format of the ground truth trajectories, which is an ASCII text file with a time stamp, a 3D position and an orientation represented as a unit quaternion. The ground truth 3D point cloud is provided as full resolution FARO raw data and subsampled in the Polygon File Format (binary PLY). 

The 2D ground truth map is extracted from the FARO data is a png with free space in white and occupied cells in black. The occupied points from the FARO data are sampled at a height of 61cm $\pm$10cm above the ground, which is the height of the horizontal Velodyne (and the height of the FARO scans), and thus the height producing 2D ground truth maps most similar to the ones from the horizontal Velodyne Lidar.

The datasets are available online \footnote{\url{https://robotics.shanghaitech.edu.cn/datasets/MARS-Dataset}}. We also provide a very short and small sample dataset from within MARS-8.

\section{Evaluation}
\label{sec:eval}

The datasets collected allow a multitude of experimentation. One can test different SLAM algorithms, each with different parameter settings. The input to those algorithms can be varied in many different ways: selection of sensors (the 9 cameras alone allow 73 different permutations), downsampling of the sensor resolution to simulate lower-quality sensors, reduction of the frame rate, for lasers also reduction of the maximum range, etc.  It is easily possible to make tens of thousands different mapping runs with the dataset (e.g. 10 algorithms, each with 5 configurations, 20 sensor combinations, 4 resolutions, 4 frame rates, 3 datasets = 48,000 maps). This is out of the scope of this paper and our future work. Instead, here we showcase a few examples using 2D laser, 3D laser (also sub-sampled), monocular and stereo SLAM, to demonstrate the principal opportunities this dataset offers.

Scientific results should be reproducible. Since we provide the dataset, we also want to give the reader the possibility to re-create the exact same maps (barring differences caused by randomized SLAM algorithms). We are thus providing ROS launch files (start scripts) that generate the maps and other needed information (e.g. the path estimated by the SLAM algorithm). We also provide the ROS package to subsample the Velodyne data as well as the ROS description package of the mapping robot, including the urdf file containing the calibration results. Together with the also provided ground truth path data it is then very easy to reproduce our evaluations, using the code from  \cite{zhang2018tutorial}.

We apply several mapping methods to our dataset, using Lidars and monocular and stereo cameras as input:
\begin{itemize}
	\item 2D Grid Mapping (converting the horizontal Velodyne scan in a 2D LRF message; 5cm resolution maps):
	\begin{itemize}
		\item Hector Mapping \cite{KohlbrecherMeyerStrykKlingaufFlexibleSlamSystem2011}
		\item Cartographer \cite{Cartographer} 
	\end{itemize} 
	\item 3D Pointcloud Mapping (with horizontal Velodyne 32 beams, also subsampled to 16 and 8 beams):
	\begin{itemize}
		\item BLAM \footnote{To support colored point clouds minor changes in the code were needed. The code is thus available on the dataset webpage. Original: \url{https://github.com/erik-nelson/blam}}
	\end{itemize}
	\item visual SLAM:
	\begin{itemize}
		\item ORB2 \cite{murTRO2015}
		\item ORB2-stereo \cite{murORB2}
	\end{itemize}
\end{itemize}

The ground truth sources for evaluation are:
\begin{itemize}
  \item Robot pose data from the tracking system (partial coverage for MARS-Loop and MARS-NoLoop).
  \item 3D pointcloud from FARO (one map, no wall for NoLoop)
  \item 2D grid map from FARO (one map, no wall for NoLoop)
\end{itemize}

The trajectory estimated by the SLAM algorithms is then compared to the trajectory of the tracking system. MARS-8 is completely covered in the tracking system, but MARS-Loop and MARS-NoLoop are only partially covered. Only the parts of the trajectories that have according ground truth poses from the tracking system are used for evaluation. The matching of those trajectories is based on the hardware-synchronized time stamps. The tracking system reported an average pose error of less than 1.5mm, which is about two orders of magnitude smaller than the SLAM estimates (see Figures \ref{fig:comparison}, \ref{fig:revocomparison} and \ref{fig:multi-comparison}). We are thus confident that the use of the tracking system allowed a proper evaluation of the trajectories estimated by the SLAM algorithms.

\begin{figure*}[tb]
    \centering
    \includegraphics[width=0.8\linewidth]{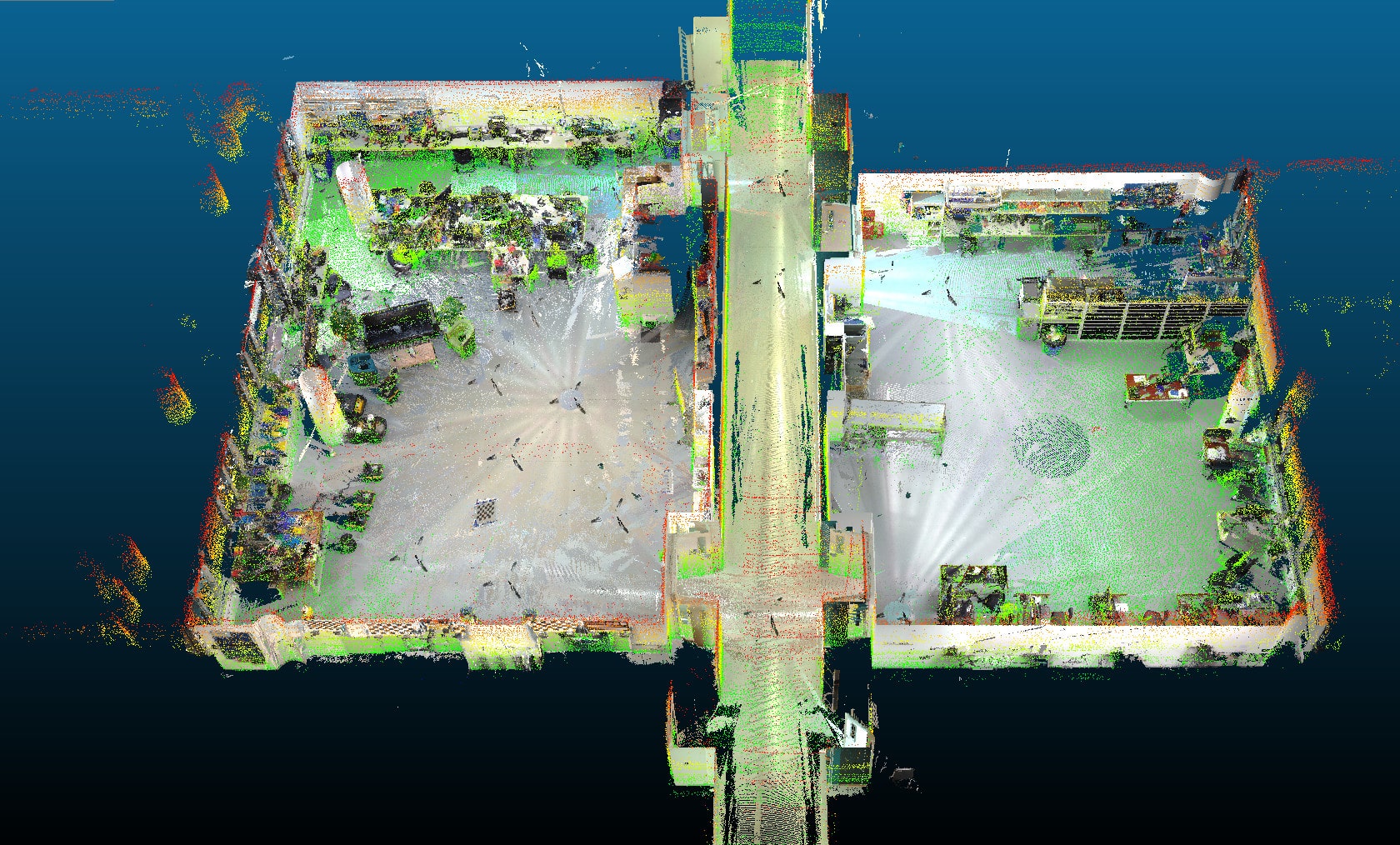}  
    \caption{Faro ground truth pointcloud overlaid with BLAM MARS-Loop in cloudcompare. In order to show the details of our datasets, we cut the ceiling.}
    \label{fig:cc}
\end{figure*}

\begin{figure*}[t]
    \centering
      \includegraphics[width=\linewidth]{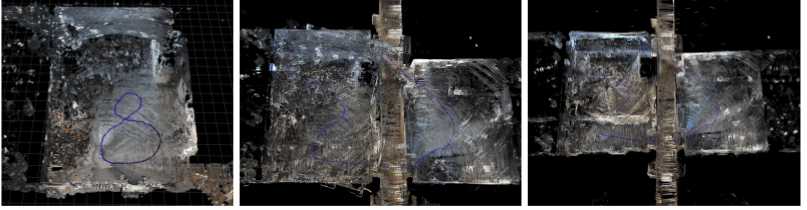}  
%
%
    
        \caption{3D colored maps created by BLAM using the horizontal 3D Lidar and the horizontal cameras.  }
        \label{fig:colorBLAM}

\end{figure*}

To quantify the quality of trajectories we get, the mean square error (RMSE)  is used to evaluate the absolute error after aligning every trajectory to its corresponding ground-truth path. The absolute translation errors are calculated to compare the performance of different algorithms. For all the above SLAM algorithms we used, all the estimated trajectories are saved. Once we get these trajectories, we estimate the performance of SLAM algorithms by using  \cite{zhang2018tutorial}'s approach for the evaluation.

%
%
%

%
%

\newcommand\cmpreSize{0.31}

\begin{figure*}[tb]
	\centering
	\parbox{\cmpreSize\linewidth}{\centering MARS-8} 
	\parbox{\cmpreSize\linewidth}{\centering MARS-Loop} 
	\parbox{\cmpreSize\linewidth}{\centering MARS-NoLoop} \\ \vspace{1.5mm}

	\rotatebox{90}{\hspace{0.3cm}Hector} 
	\includegraphics[width=\cmpreSize\linewidth]{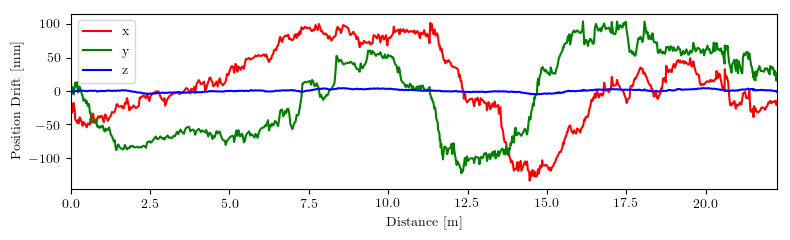}  
	\includegraphics[width=\cmpreSize\linewidth]{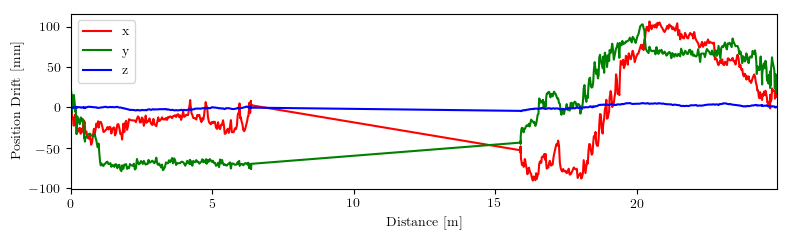}  
	\includegraphics[width=\cmpreSize\linewidth]{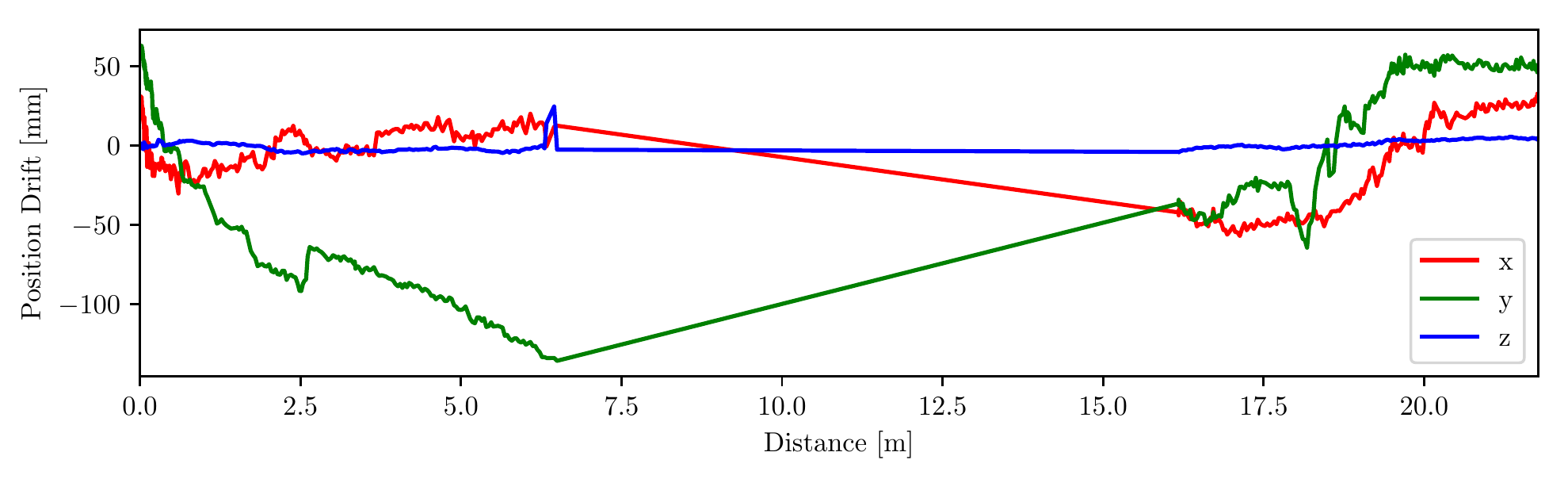}  \\ \vspace{1.5mm}
	
	\rotatebox{90}{\hspace{0.7cm}Cart.} 
	\includegraphics[width=\cmpreSize\linewidth]{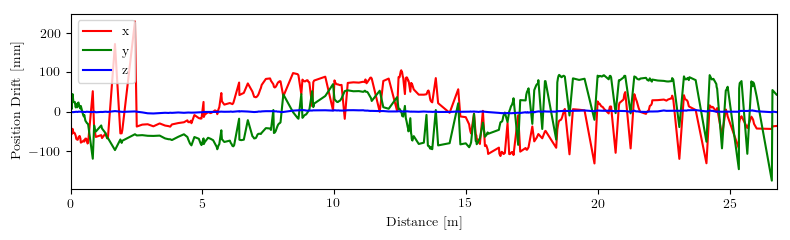}  
	\includegraphics[width=\cmpreSize\linewidth]{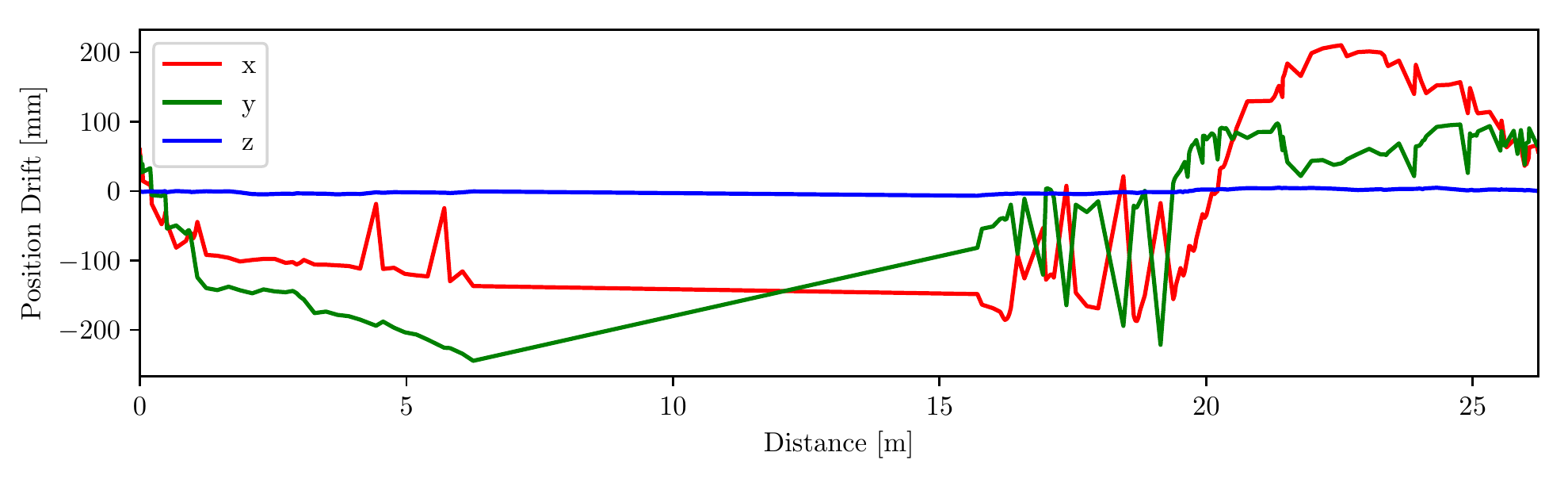}  
	\includegraphics[width=\cmpreSize\linewidth]{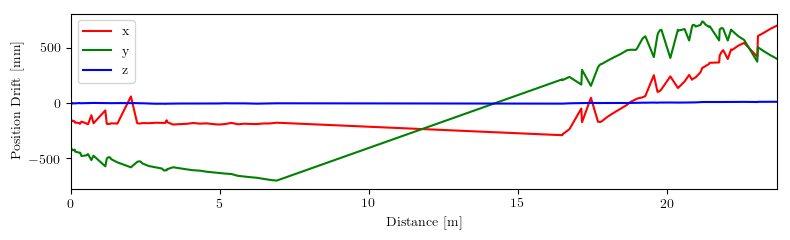}   \\ \vspace{1.5mm}
	
	\rotatebox{90}{\hspace{0.1cm}BLAM-32} 
    \includegraphics[width=\cmpreSize\linewidth]{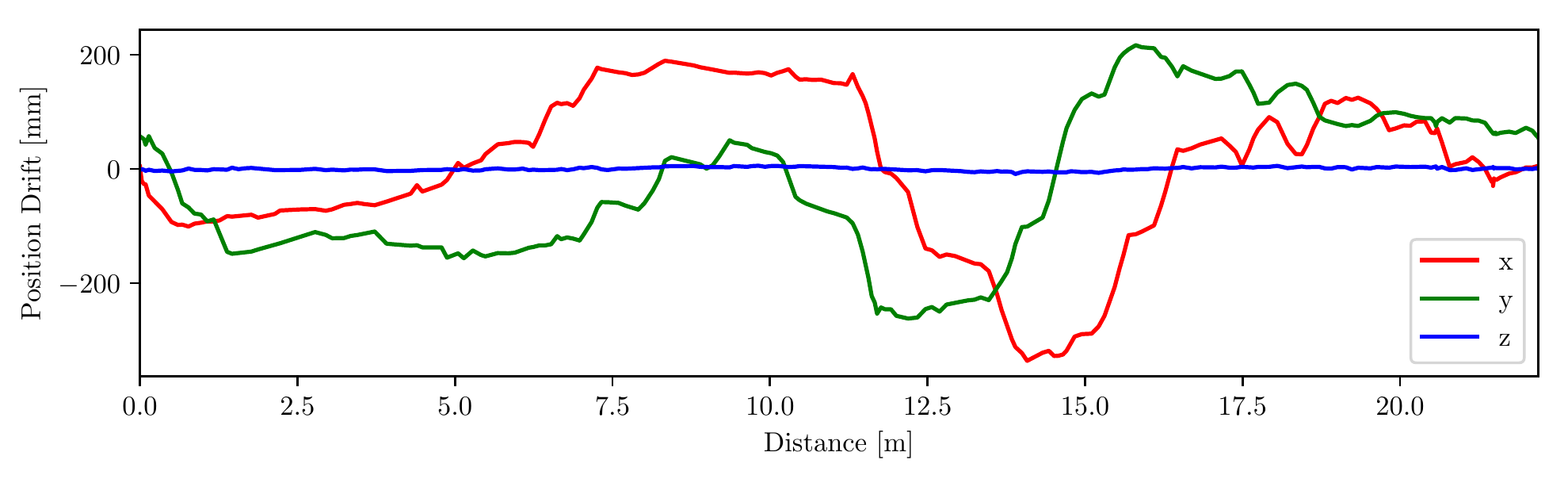}  
    \includegraphics[width=\cmpreSize\linewidth]{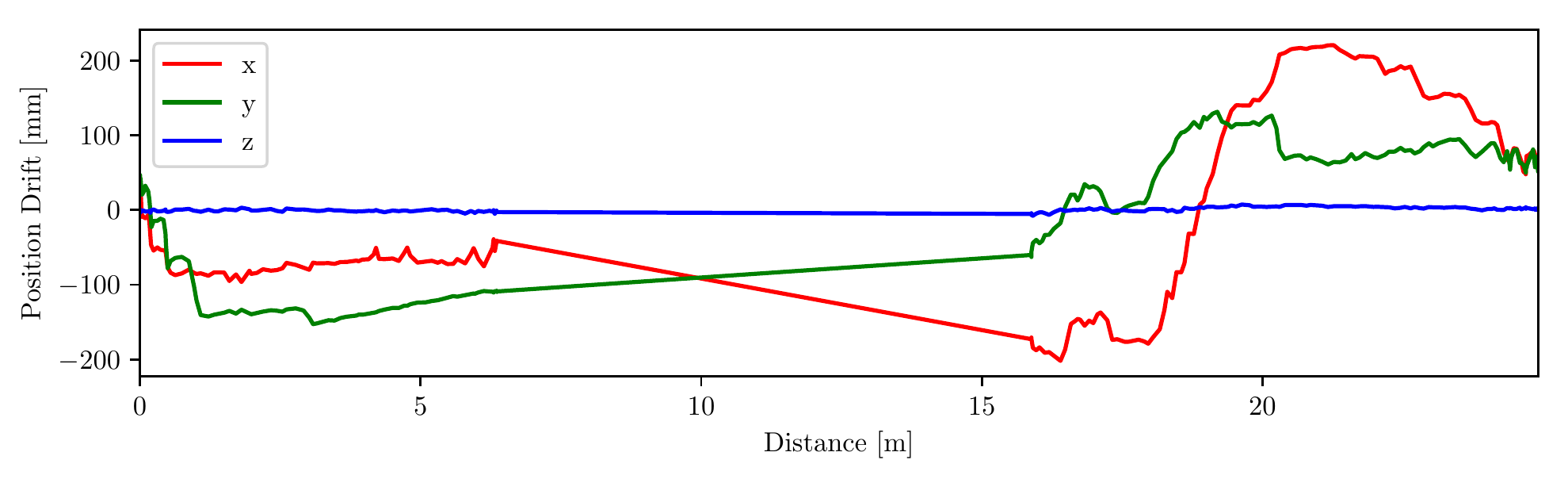}  
    \includegraphics[width=\cmpreSize\linewidth]{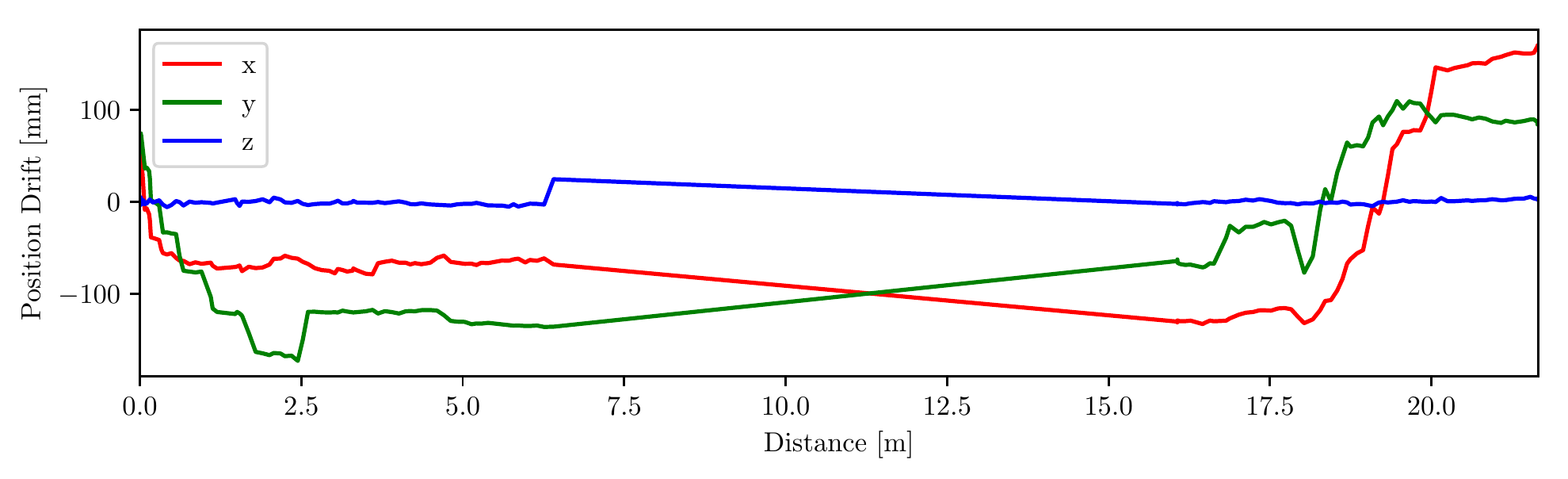}   \\ \vspace{1.5mm}

	\rotatebox{90}{\hspace{0.5cm}ORB2} 
	\includegraphics[width=\cmpreSize\linewidth]{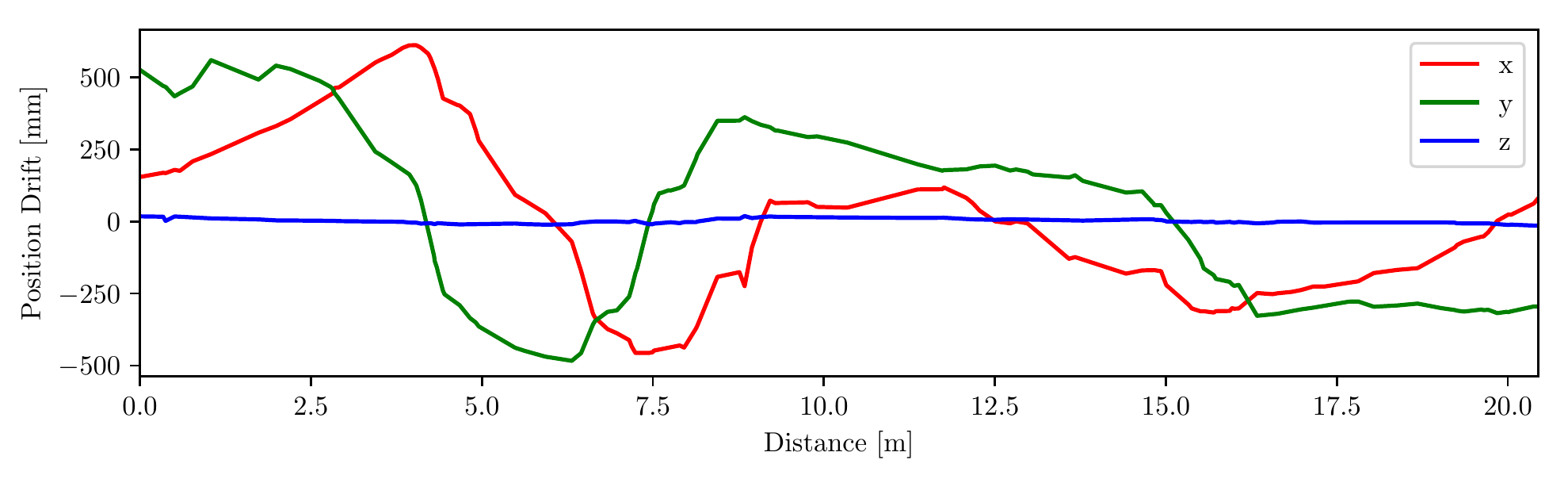}  
	\includegraphics[width=\cmpreSize\linewidth]{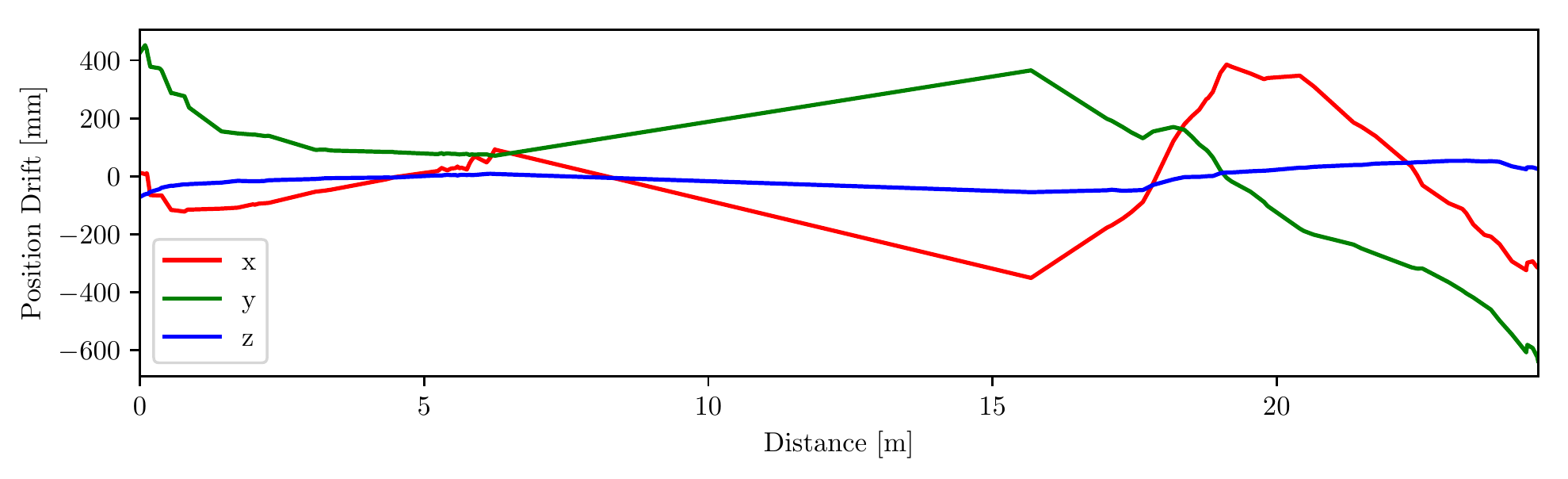} 
	\parbox[b][1.8cm][c]{\cmpreSize\linewidth}{\centering Failed} \\ \vspace{1.5mm}

	\rotatebox{90}{\hspace{0.1cm}ORB2 St.} 
	\includegraphics[width=\cmpreSize\linewidth]{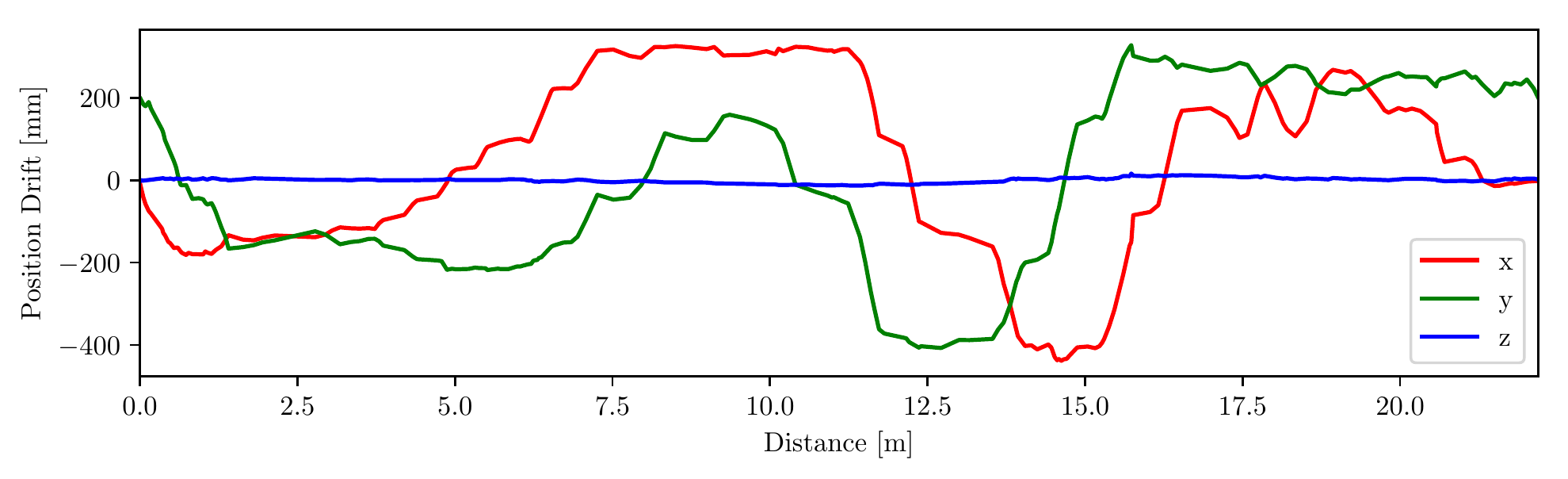} 
	\includegraphics[width=\cmpreSize\linewidth]{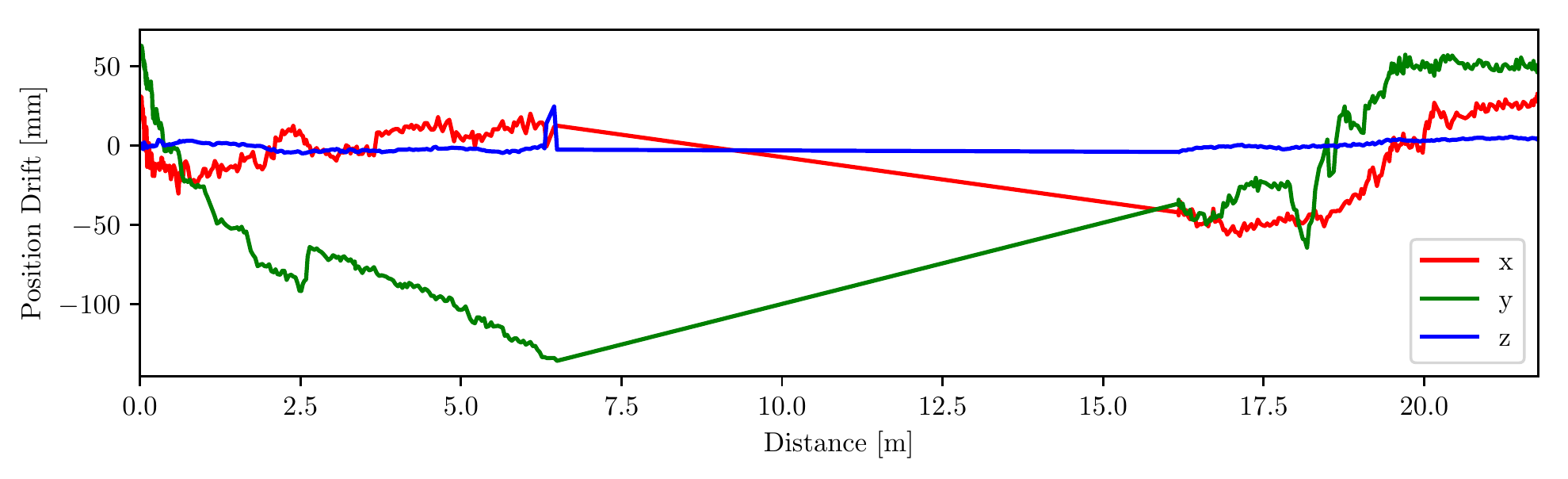} 
	\parbox[b][1.8cm][c]{\cmpreSize\linewidth}{\centering Failed} \\ \vspace{1mm} 
	
	\caption{The translation error in all three axes for the part of the trajectories covered by the tracking system. On the top Hector Mapping with MARS-8 (left), MARS-Loop (middle) and MARS-NoLoop (right). Below cartographer, BLAM, ORB2 and ORB2-Stereo is at the bottom.}
	\label{fig:comparison}
	
\end{figure*}

\begin{figure*}[tb]
	\centering
	\parbox{\cmpreSize\linewidth}{\centering MARS-8} 
	\parbox{\cmpreSize\linewidth}{\centering MARS-Loop} 
	\parbox{\cmpreSize\linewidth}{\centering MARS-NoLoop} \\ \vspace{1.5mm}

	\rotatebox{90}{\hspace{0.3cm}BLAM-8} 
	\includegraphics[width=\cmpreSize\linewidth]{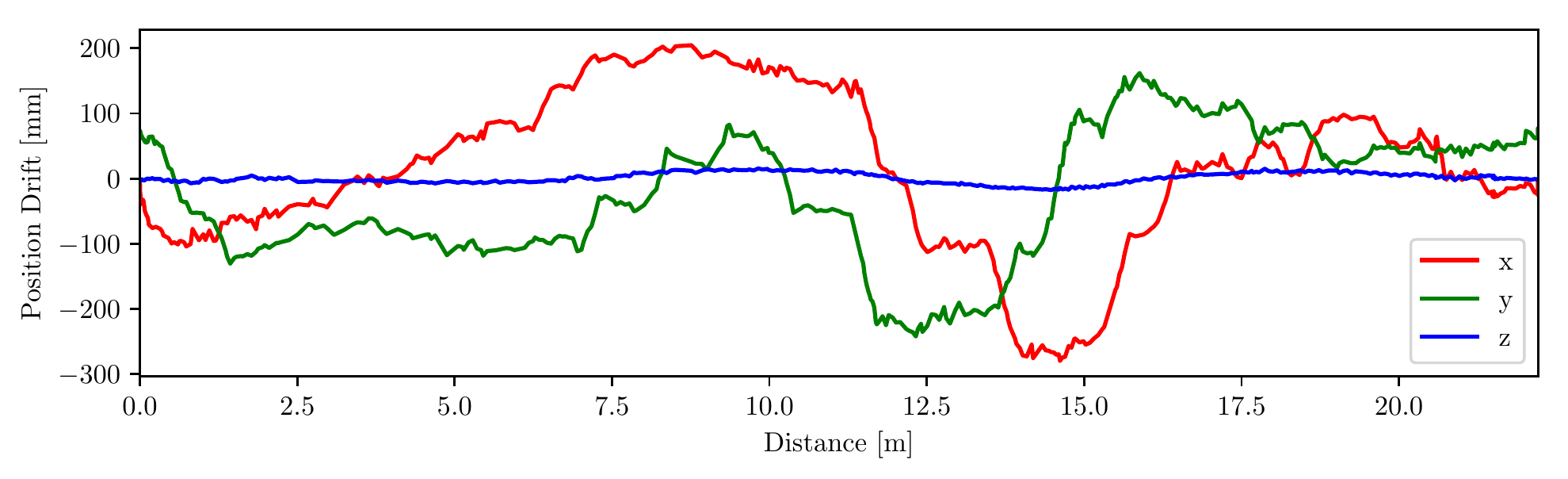}  
	\includegraphics[width=\cmpreSize\linewidth]{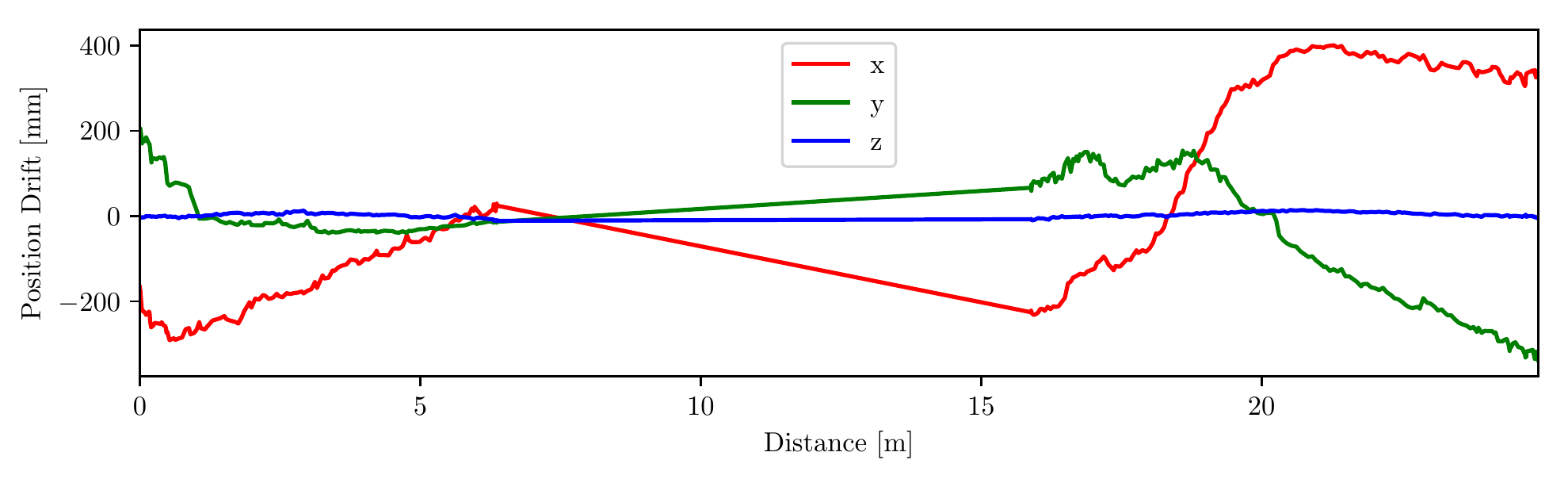}  
	\includegraphics[width=\cmpreSize\linewidth]{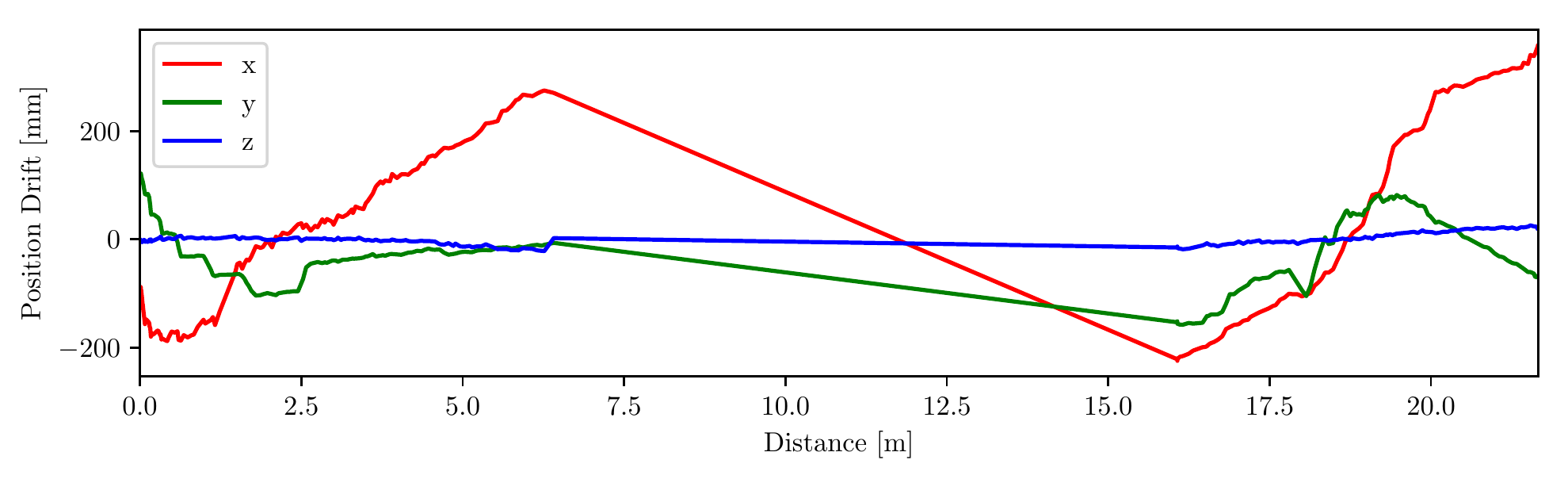}  \\ \vspace{1.5mm}

	\rotatebox{90}{\hspace{0.3cm}BLAM-16} 
	\includegraphics[width=\cmpreSize\linewidth]{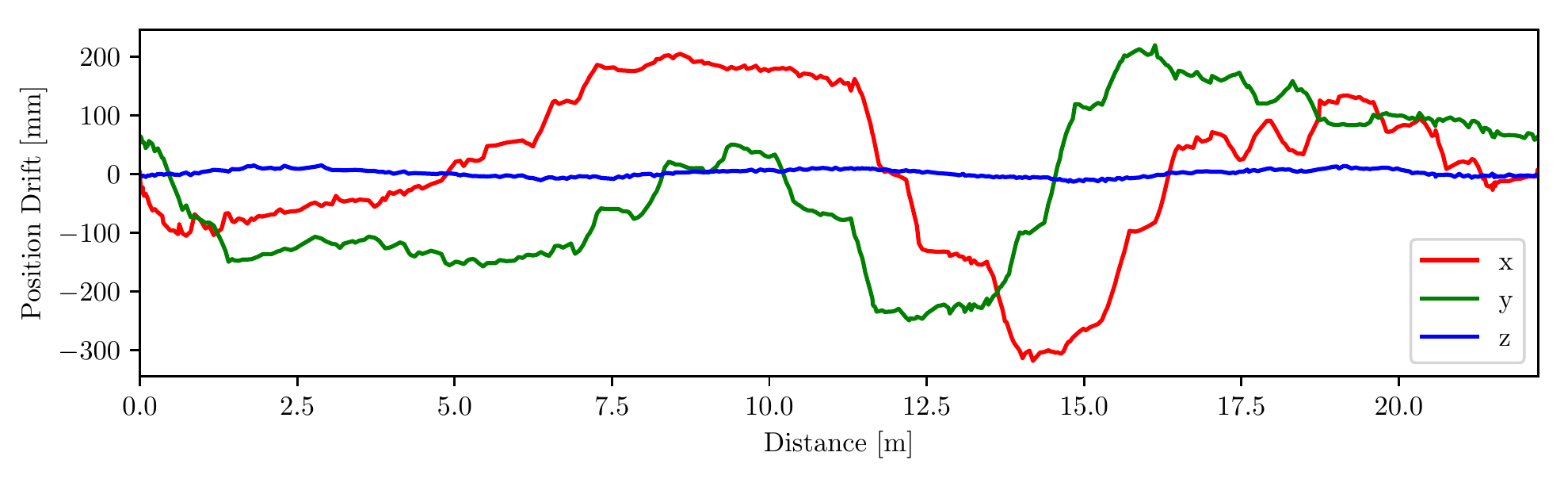}  
	\includegraphics[width=\cmpreSize\linewidth]{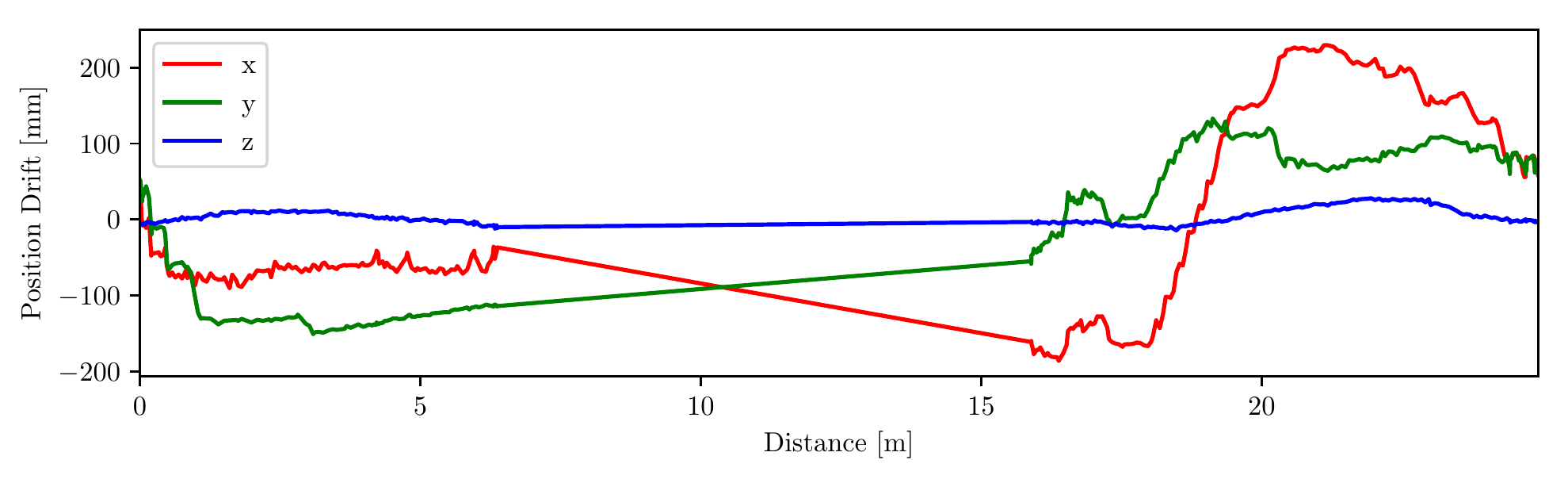}  
	\includegraphics[width=\cmpreSize\linewidth]{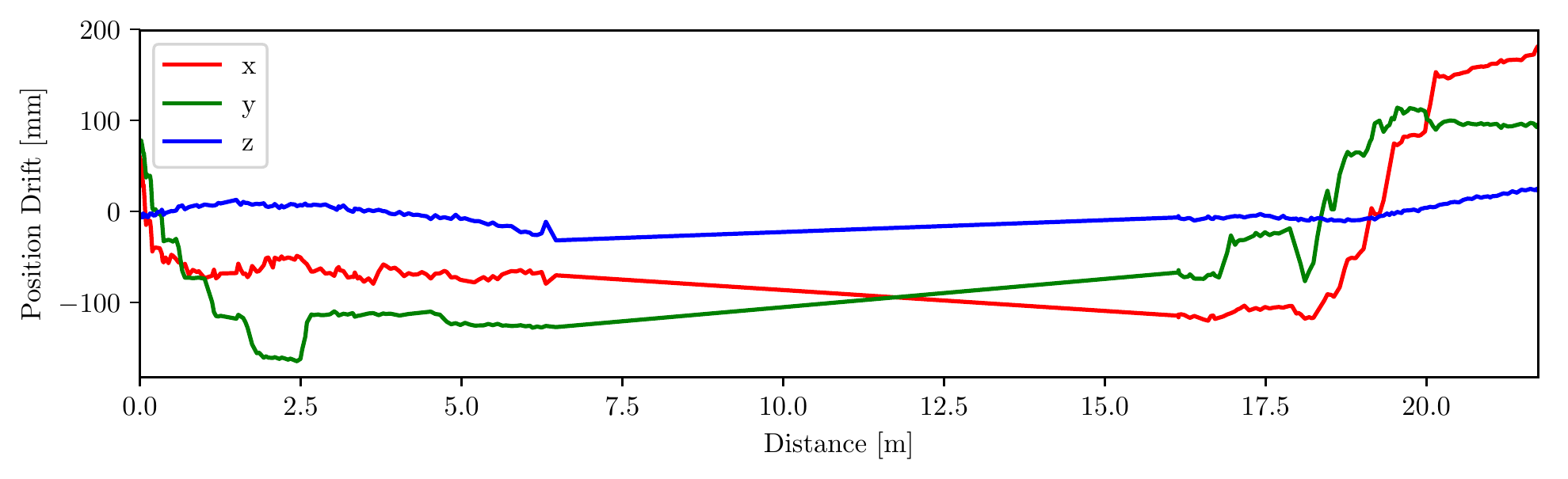}   \\ \vspace{1.5mm}

	\rotatebox{90}{\hspace{0.3cm}BLAM-32} 
	\includegraphics[width=\cmpreSize\linewidth]{sections/revision-result/8-32.pdf}  
	\includegraphics[width=\cmpreSize\linewidth]{sections/revision-result/loop32.pdf}  
	\includegraphics[width=\cmpreSize\linewidth]{sections/revision-result/noloop32.pdf}   \\ \vspace{1.5mm}

	\caption{The translation error in all three axes for the part of the trajectories covered by the tracking system. On the top Mapping with MARS-8 (left), MARS-Loop (middle) and MARS-NoLoop (right). BLAM-8 means we select every 4th beam, BLAM-16 means we select every 2nd beam, BLAM-32 means we select all 32 beams for mapping. }
	\label{fig:revocomparison}
	
\end{figure*}

From Figure\ref{fig:revocomparison}, we can see that for the MARS-8 dataset, by selecting every 4th, 2nd and full beams from Velodyne beams, the translation error does not have a significant difference. But for MARS-LOOP and MARS-NoLoop dataset, the error of the x-axis for BLAM-8 is a bit larger than BLAM-16 and BLAM-32.  The error of the x-axis in BLAM-8 and BLAM-16 has no significant difference.

\begin{figure*}[tb]
	\centering
	\parbox{\cmpreSize\linewidth}{\centering MARS-8} 
	\parbox{\cmpreSize\linewidth}{\centering MARS-Loop} 
	\parbox{\cmpreSize\linewidth}{\centering MARS-NoLoop} \\ \vspace{1.5mm}
	
	\rotatebox{90}{\hspace{0.8cm}BLAM} 
	\includegraphics[width=\cmpreSize\linewidth]{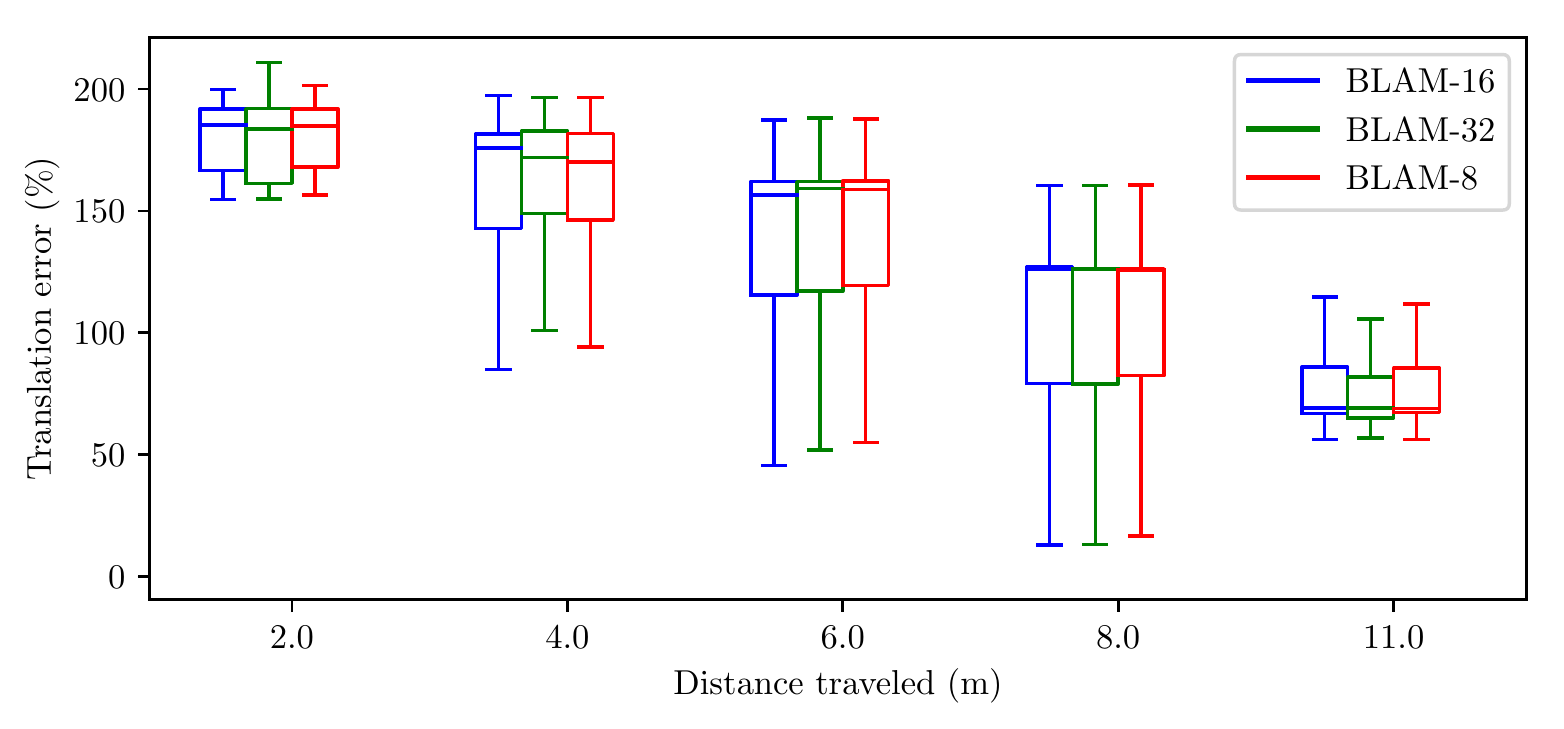}  
	\includegraphics[width=\cmpreSize\linewidth]{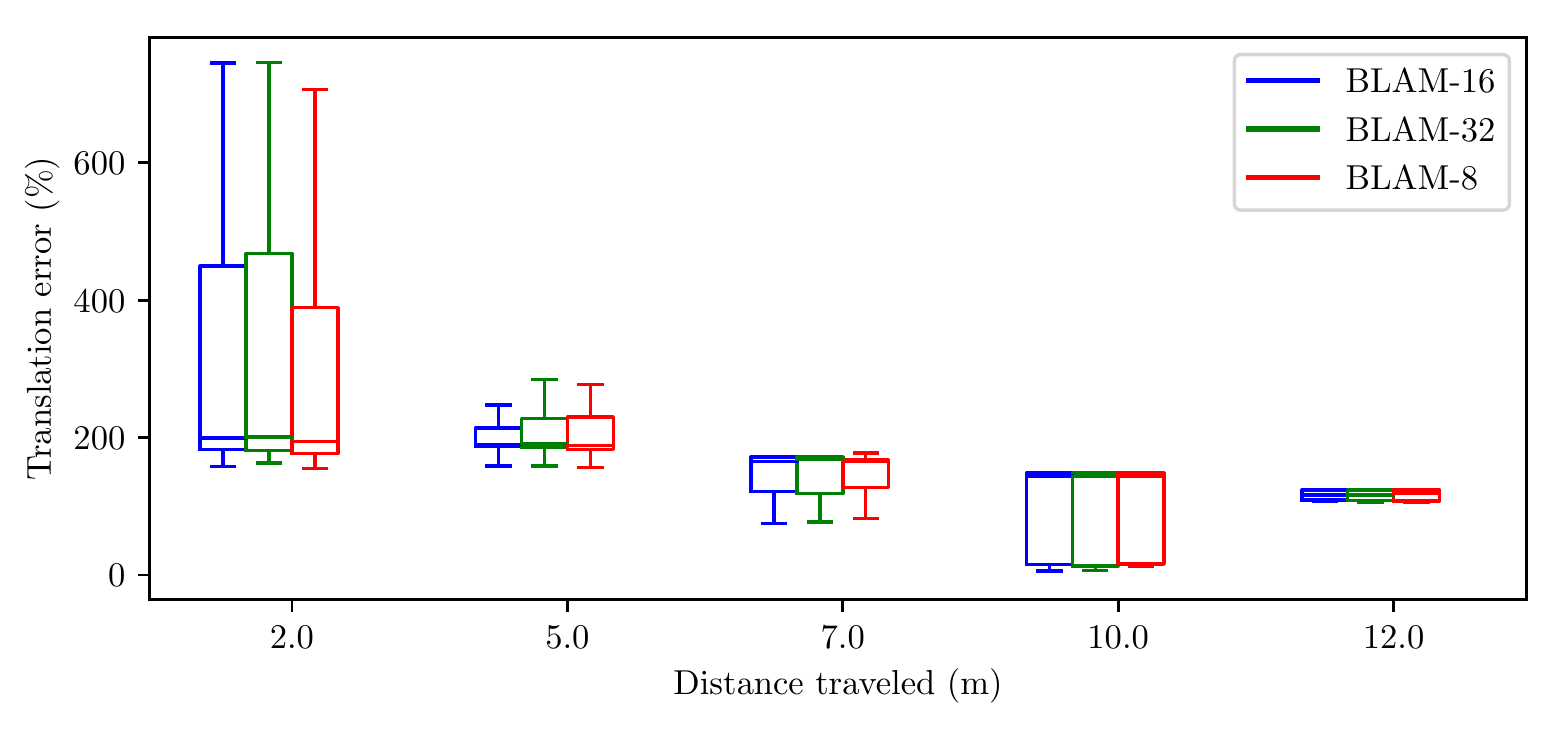}  
	\includegraphics[width=\cmpreSize\linewidth]{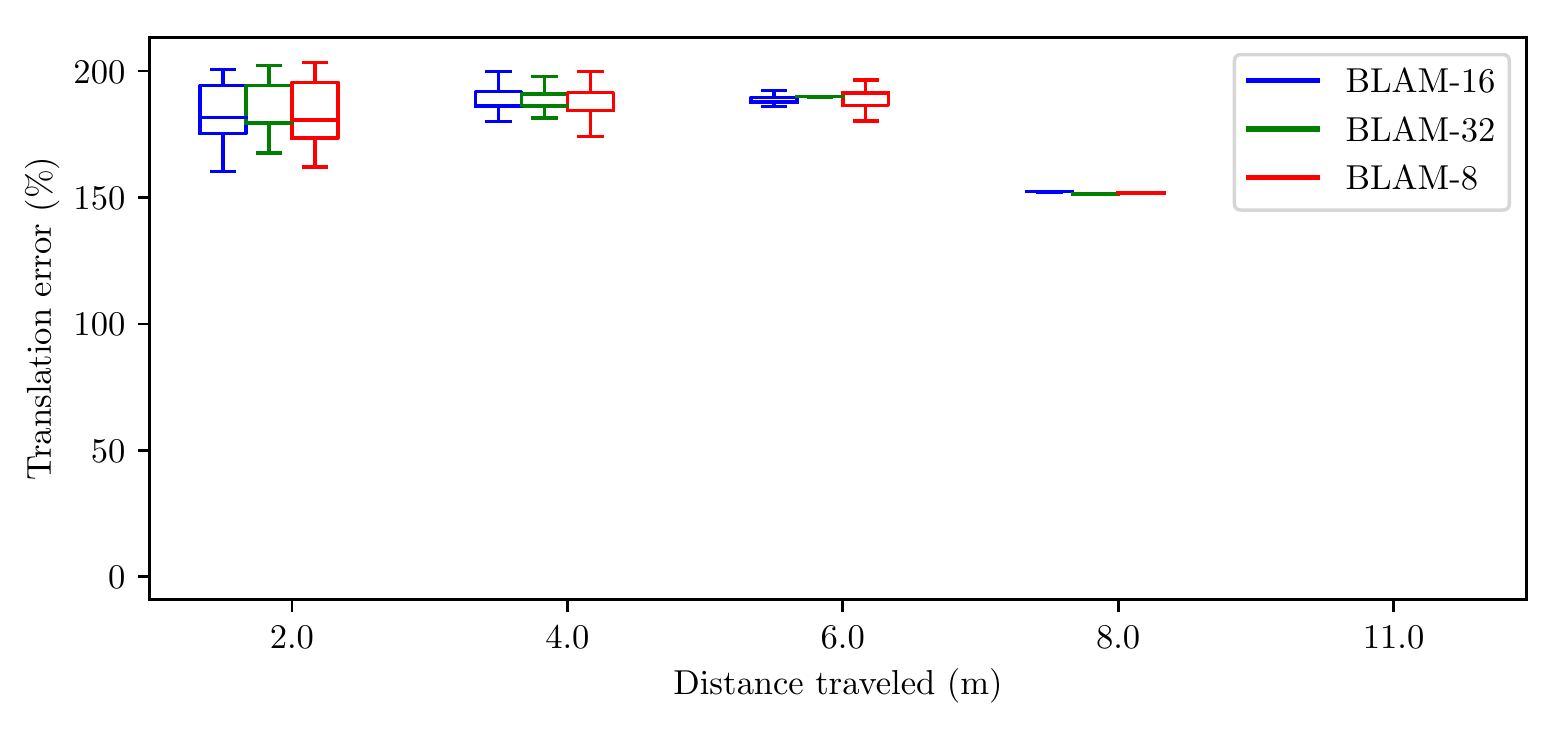}   \\ \vspace{1.5mm}
	
	\caption{The translation errors of BLAM. BLAM-8 means we select every 4th beam, BLAM-16 means we select every 2nd beam, BLAM-32 means we select all 32 beams for mapping. On the top mapping with MARS-8 (left), MARS-Loop (middle) and MARS-NoLoop (right).}
	\label{fig:multi-comparison}
	
\end{figure*}

Figure \ref{fig:multi-comparison} shows the box plot results of BLAM by using different numbers of Velodyne beams.  We can see that there aren't many differences when using 8 beams, 16 beams or a full scan of Velodyne beams.
For the BLAM mapping algorithm and our dataset, we found that fewer beams could have almost the same performance for mapping. For future work it will be interesting if we can collect datasets that result in different performances for the subsampled BLAM versions.

\begin{figure*}[h!]
	\includegraphics[width=\linewidth]{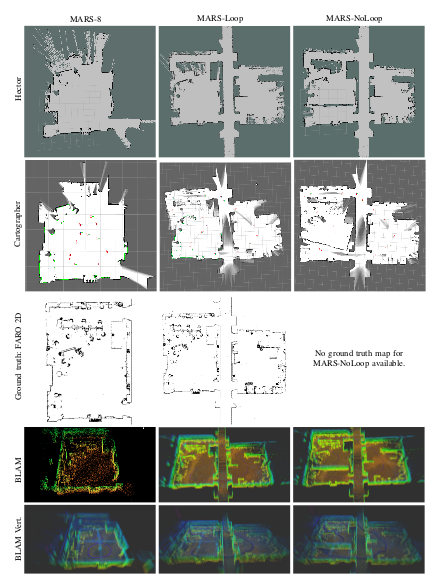}    \\ \vspace{1mm}	

	\caption{ The top three rows are 2D grid maps created by Hector Mapping on top, cartographer middle and Ground Truth FARO 2D   on maps MARS-8, MARS-Loop and MARS-NoLoop, respectively.  In the bottom two rows are 3D maps created by BLAM using our horizontal and vertical 3D Lidar on MARS-8, MARS-Loop and MARS-NoLoop, respectively.}
	\label{fig:maps_2d}
\end{figure*}

\begin{figure*}[h!]
	\includegraphics[width=0.8\linewidth]{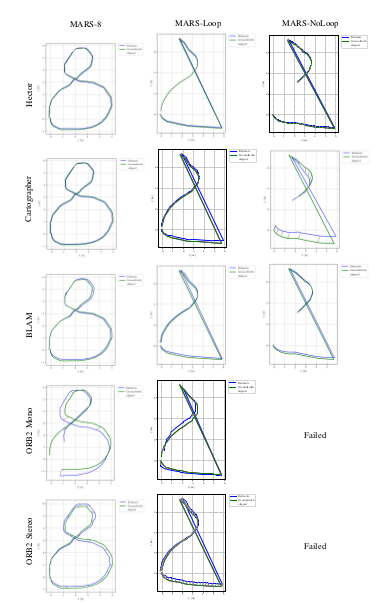}  
	\centering
	\caption{Plots of the paths estimated by the SLAM algorithms matched against the ground truth paths.}
	\label{fig:path_eval}
\end{figure*}

Figure \ref{fig:maps_2d} shows the 2D grid maps from Hector Mapping, Cartographer, Ground Truth FARO 2D and 3D point clouds from BLAM. We see that cartographer  has problems when coming back into the MARS lab (on the left) and no loop closing is possible. It also has a broken map in MARS-Loop. Figure \ref{fig:comparison} shows the error in the trajectories in all three axes by Hector (top), Cartographer, BLAM, ORB2 and ORB2-Stereo. For MARS-Loop and MARS-NoLoop note the jump in the data in the middle: This is where the robot left the tracking system and later re-entered it. This jump can also be seen in Figure \ref{fig:path_eval}.

\newcommand\blaSize{0.32}

\begin{figure*}[t]
       \includegraphics[width=\linewidth]{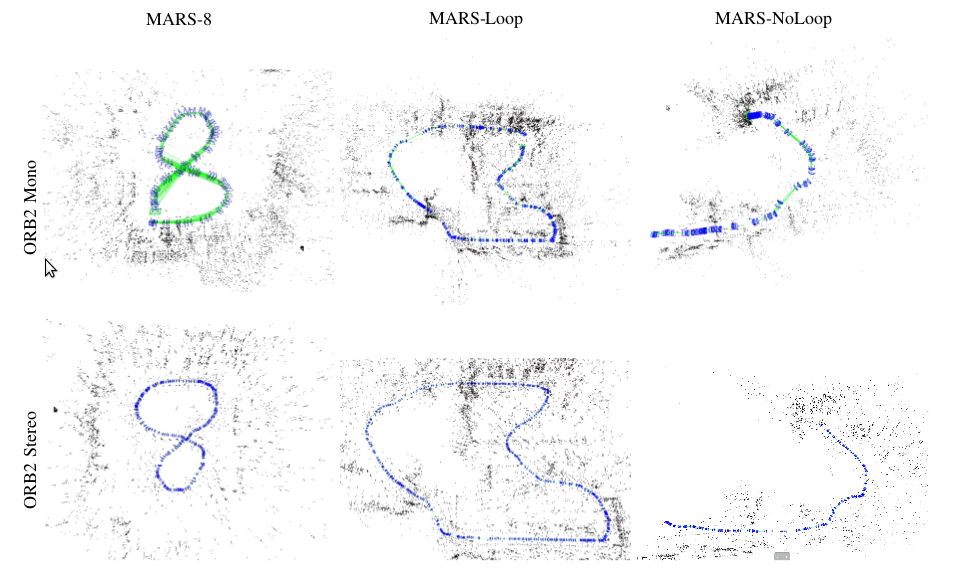}  
    \centering
%
%
%
    
    
        \caption{
                The feature clouds and trajectories  created by ORB2 and ORB2-stereo for MARS-8, MARS-Loop and MARS-NoLoop, respectively. For the MARS-NoLoop dataset ORB2 and ORB2-stereo fail to create whole trajectories.}

        \label{fig:orb_error}

\end{figure*}

 We can see that the shown error correlates nicely with the perceived map quality of Figure \ref{fig:maps_2d}. Figure \ref{fig:comparison} quantifies the error  in a diagram. It shows absolute errors of 10cm for Hector MARS-8, MARS-Loop and MARS-NoLoop. The results of Hector are quite good. For BLAM among all our datasets we can see that absolute errors of $x$ axis and $y$ axis are about 20cm. We can make use of the trajectory evaluation shown in Figure \ref{fig:comparison}. We see that the error is low, but looking at Figure \ref{fig:comparison} we see that it is double the value of the good Hector maps. The BLAM pointclouds in Figure \ref{fig:maps_2d} are good and nicely show the double curtain in MARS-NoLoop. 

 The ORB2 algorithm is using just one camera (forward-looking on the left side), while the ORB2-Stereo SLAM algorithm is using both front cameras for stereo. The feature clouds and trajectories are shown in Figure \ref{fig:orb_error}. 
 Both visual SLAM approaches have problems when coming back into the MARS lab and no loop closing is possible. Both of them fail to create a complete trajectory on MARS-NoLoop dataset. On the one hand this is caused by the sudden change in appearance when driving through the door. But on the other hand, the fact that no loop closing is possible is of course making it impossible to correct the localization, as it most likely happened in MARS-Loop.

In the bottom of Figure \ref{fig:maps_2d} the 3D maps generated with BLAM by using the horizontal Velodyne and vertical Velodyne are shown. First the SLAM result of BLAM with the horizontally scanning sensor are shown. For comparison, the last row of Figure \ref{fig:maps_2d}  is generated using the localization estimate from the horizontal BLAM while taking the point clouds from the vertical scanner for mapping. It can be seen that those maps show more details on the ceiling (e.g. the truss for the tracking system) while providing a similar level of detail on the walls. The map from the vertical scanner is lacking the points along the robot path for MARS-Loop and MARS-NoLoop, because the robot is not driving big turns within the same room and thus not scanning the previous robot positions.


We have also employed cloudcompare\footnote{\url{http://cloudcompare.org/}} for quality measurement. We register the Faro pointcloud with the robot pointcloud from BLAM-Loop and then calculate the RMS. The result is an RMS of 0.084 with a theoretical overlap of 90\%. Figure \ref{fig:cc} shows the two pointclouds overlaid. We didn't collect a Faro pointcloud for BLAM-NoLoop. 

For the last row of \ref{fig:colorBLAM} we use the colored pointcloud to build 3D maps with color. After calibrating all sensors mounted on our mapping robot, the relative poses of cameras and 3D Lidar are known, so we can color the pointcloud by RGB images by using ray-tracing. We use all cameras except the two upward facing cameras to color the pointcloud of the horizontally scanning Velodyne. BLAM is then using the colored pointclouds to build the 3D colored maps.


\renewcommand\blaSize{0.30}

\newcommand\pathSize{0.25}

\section{Conclusions}
\label{sec:conclusions}
Mapping datasets are essential for comparing SLAM algorithms and, as our survey revealed, up to now there is no publicly available dataset satisfying the requirements of high-resolution, multi-sensor, hardware-synchronized data with ground-truth path and map information. To the best of our knowledge, we have created one of the most performant mapping dataset collection robots world-wide, featuring 5 megapixel stereo-camera pairs to the front, left, right and top and a monocular camera to the back, as well as two 32-beam Lidar scanners and an IMU. The sensors are all hardware-synchronized and calibrated to the robot base frame. 

This data enables the comparison of various mapping approaches (e.g. front and back stereo, ceiling mapping, fused Lidar and monocular, etc.) with different SLAM software. We could also compare visual SLAM with different resolutions by down-scaling the images or 2D SLAM by extracting 2D Lidar scans from the 3D Lidar sensors. The ground truth path and map information allows for the proper evaluation of the SLAM runs. We believe that this data is very valuable for SLAM researchers. The amount of data and resolution of the images might seem excessive, but future robotic systems quite likely will feature such rich data, since sensor quality and processing power are growing exponentially while the costs are steadily decreasing. Granted, with about 170 MB/s our datasets are pretty big, but this is unavoidable with the presented requirements. 

The three indoor datasets we collected cover a relatively small area. MARS-8 is covered in the tracking system, while MARS-Loop and MARS-NoLoop leave and later re-enter the room and the tracking system. Even though they are not long, the MARS-Loop and MARS-NoLoop datasets already prove to be challenging for some very well known mapping algorithms evaluated in this paper. We were running 2D Lidar SLAM software, 3D Lidar SLAM, and a monocular and a stereo visual SLAM approach. The results confirm the intuition that, using loop closures, the error of maps can be reduced. Furthermore, we showed that, at least for our dataset and the algorithms selected, the 3D Lidar approach gives the most accurate results. The details of SLAM datasets we surveyed int Section \ref{sec:dataset_survey} are available online  \footnote{\url{https://robotics.shanghaitech.edu.cn/datasets/MARS-Dataset}}. We also provide a video to show the details of our MARS Mapper　Robot. It can also be found in the same website.

In the future we plan to upgrade the mapping hardware. We want to collect the images with 60Hz, include a back-stereo setup, use more and higher resolution Lidars and add other types of sensor information (e.g. sonar; event-, infrared- and panoramic- cameras; dGPS for outdoors). This will require a bigger robot with much more processing power and unfortunately also generate much bigger datasets, but at the same time we will be able to test SLAM algorithms we even more diverse setups. For example will we be able to explore visual SLAM with different resolutions, frame-rates and camera-configurations. We then also plan to collect bigger (area covered) and more diverse datasets. This will then enable us to evaluate SLAM algorithms in hundreds of different configurations.

\nocite{*}


\IEEEtriggeratref{83}

\bibliographystyle{IEEEtran}
\bibliography{slamdataset_chy}
%


\onecolumn

\appendices
\section{}
\label{sec:appendix}
The following table is an extract from an excel file we created to capture various information about robotics-related datasets. The excel file is available as an attachment to this publication and additionally contains some more data and links to the according websites. Please see the index on the next page for the abbreviations used in the table. \\ \\


 \includegraphics[height=1\linewidth,angle=90]{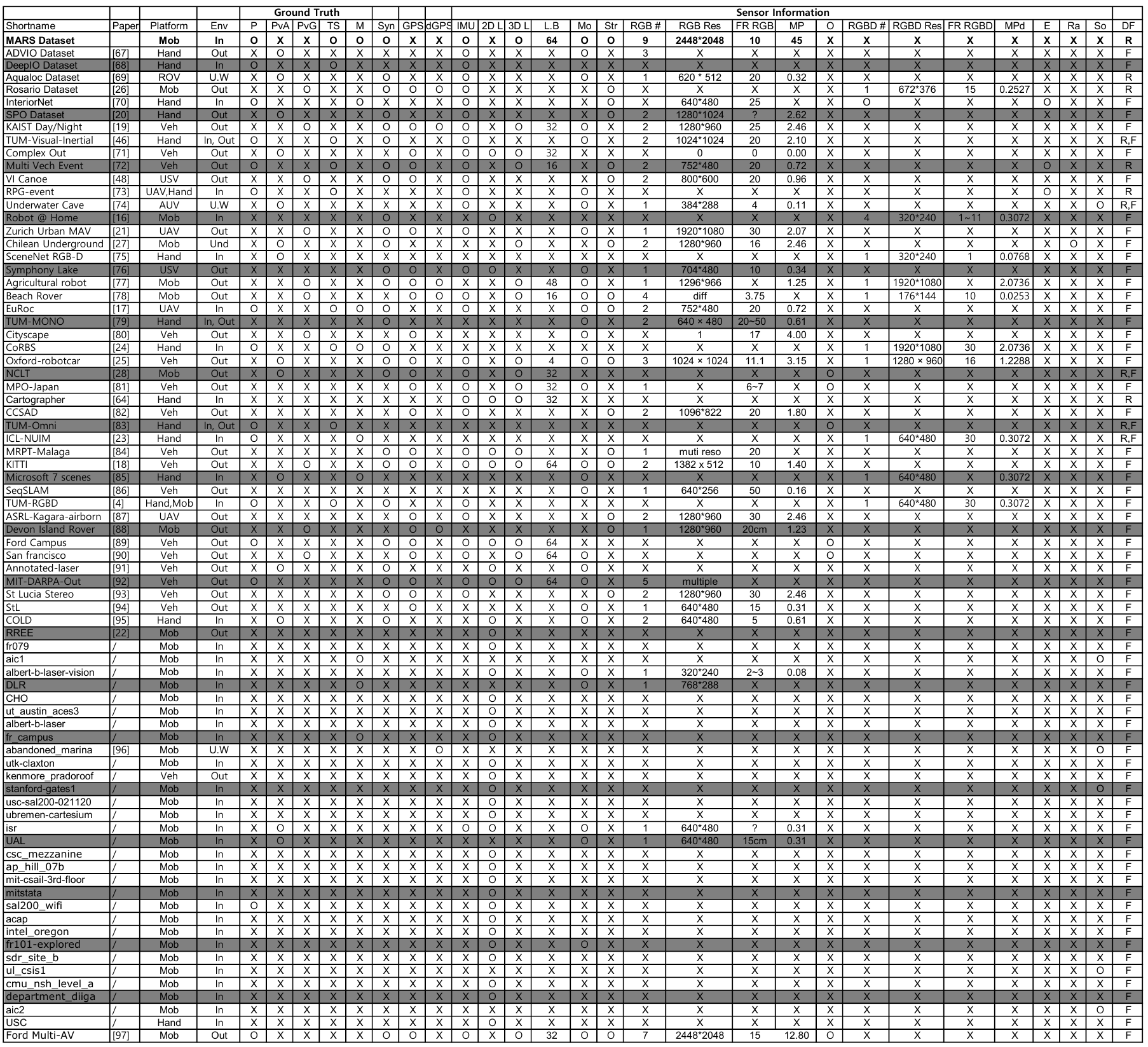}  

\vfill


\begin{figure*}[h!]
	\centering
	Abbreviation Index\\ \vspace{0.3cm}
	\includegraphics[width=8.4cm]{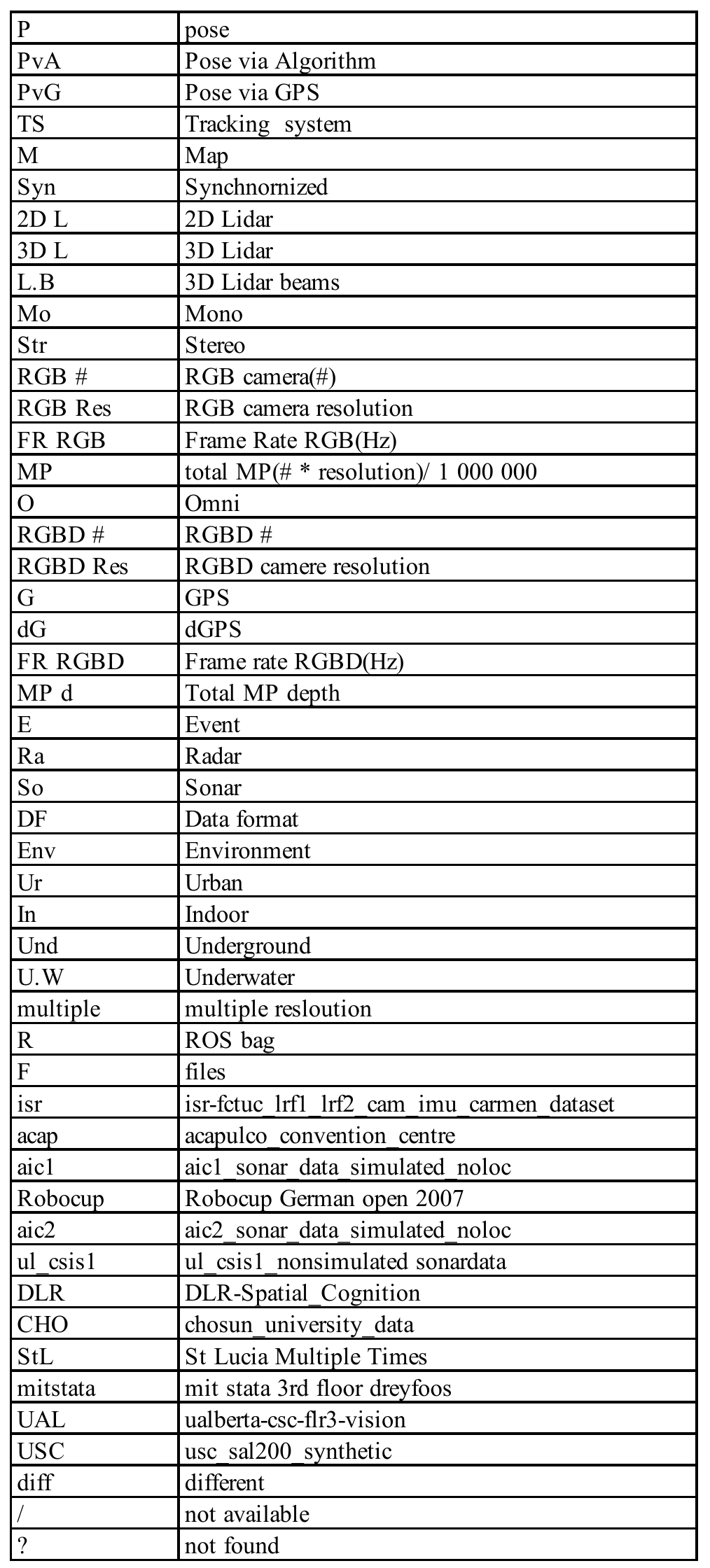}  
\end{figure*}

\vspace{5cm}

\end{document}